\documentclass{article}

\usepackage[preprint]{corl_2026} 

\usepackage{multicol}
\usepackage{amsmath}
\usepackage{amsfonts}
\usepackage{array}
\usepackage{float}
\usepackage{graphicx}
\usepackage[table]{xcolor}
\usepackage[dvipsnames,svgnames]{xcolor}
\usepackage{booktabs}
\usepackage{tabularx}
\usepackage{makecell}
\usepackage{colortbl}
\usepackage{multirow}
\usepackage{adjustbox}
\usepackage{subcaption}
\usepackage{csquotes}
\usepackage{cite}
\usepackage{cleveref}
\usepackage{paralist}
\usepackage{tikz}
\usepackage{siunitx}
\usepackage{wrapfig}
\usepackage{diagbox}

\newcommand{\mycomment}[1]{\textcolor{gray}{\(\triangleright\) #1}}
\usepackage[ruled,vlined,linesnumbered]{algorithm2e} 
\SetKwComment{tcp}{\# }{}

\definecolor{myred}{RGB}{255, 200, 200}
\definecolor{mygreen}{RGB}{200, 255, 200}
\definecolor{mygrey}{rgb}{0.92, 0.92, 0.95}

\newcommand{\splitcelltwo}[6]{%
    \begin{tikzpicture}[
        baseline=(current bounding box.center), 
        x=1.8cm, 
        y=0.5cm, 
        inner sep=0pt, 
        outer sep=0pt
    ]
        \fill[#1] (0,0) rectangle (#2,1);
        \fill[#4] (#2,0) rectangle (1,1);
        \node at (#2/2, 0.5) {\scriptsize #3}; 
        \node at ({#2 + (1-#2)/2}, 0.5) {\scriptsize #6};
    \end{tikzpicture}%
}

\newcommand{\fullcell}[2]{%
    \begin{tikzpicture}[
        baseline=(current bounding box.center), 
        x=1.8cm, 
        y=0.5cm, 
        inner sep=0pt, 
        outer sep=0pt
    ]
        \fill[#1] (0,0) rectangle (1,1);
        \node at (0.5, 0.5) {\scriptsize #2}; 
    \end{tikzpicture}%
}

\newcommand{\splitcellthree}[9]{%
    \begin{tikzpicture}[
        baseline=(current bounding box.center), 
        x=1.8cm, 
        y=0.5cm, 
        inner sep=0pt, 
        outer sep=0pt
    ]
        \fill[#1] (0,0) rectangle (#2,1);
        \fill[#4] (#2,0) rectangle (#5,1);
        \fill[#7] (#5,0) rectangle (1,1);
        
        \node at ({#2/2}, 0.5) {\scriptsize #3};
        \node at ({(#2 + #5)/2}, 0.5) {\scriptsize #6};
        \node at ({(#5 + 1)/2}, 0.5) {\scriptsize #9};
    \end{tikzpicture}%
}

\usepackage[most]{tcolorbox} 
\usepackage{xcolor}

\newtcolorbox{promptbox}{
    colback=gray!10,       
    colframe=gray!40,      
    arc=4mm,               
    boxrule=0.5mm,         
    left=10pt, right=10pt, 
    top=8pt, bottom=8pt,
    fonttitle=\bfseries,
    title=Qwen Input Prompt,
}

\definecolor{dinocolor}{RGB}{240, 128, 128}
\definecolor{vintcolor}{RGB}{160,  82,  45}
\definecolor{nomadcolor}{RGB}{128,   0, 128}
\definecolor{metnetcolor}{RGB}{255,  20, 147}
\definecolor{refcolor}{RGB}{50, 205,  50} 

\newcommand{\prettydino}{\textcolor{dinocolor}{\textbf{VISTA}}}
\newcommand{\prettyablation}{\textcolor{cyan}{\textbf{VISTA w/o AH}}}
\newcommand{\prettyvint}{\textcolor{vintcolor}{\textbf{ViNT}}}
\newcommand{\prettynomad}{\textcolor{nomadcolor}{\textbf{NoMaD}}}
\newcommand{\prettymetnet}{\textcolor{metnetcolor}{\textbf{MetricNet}}}
\newcommand{\prettyref}{\textcolor{refcolor}{\textbf{Reference}}}

\definecolor{recon_color}{RGB}{249, 44, 45}
\newcommand{\prettyrecon}{\textcolor{recon_color}{\textbf{Recon}}}

\definecolor{stanford_color}{RGB}{62, 6, 79}
\newcommand{\prettystanford}{\textcolor{stanford_color}{\textbf{Go Stanford}}}

\definecolor{tartan_color}{RGB}{249, 129, 25}
\newcommand{\prettytartan}{\textcolor{tartan_color}{\textbf{Tartan Drive}}}

\definecolor{scand_color}{RGB}{231, 35, 192}
\newcommand{\prettyscand}{\textcolor{scand_color}{\textbf{Scand}}}

\definecolor{sacson_color}{RGB}{127, 127, 127}
\newcommand{\prettysacson}{\textcolor{sacson_color}{\textbf{SACSoN}}}

\definecolor{gymnasium_color}{RGB}{86, 128, 176}
\newcommand{\prettygymnasium}{\textcolor{gymnasium_color}{\textbf{Gymnasium}}}

\definecolor{library_color}{RGB}{255, 216, 47}
\newcommand{\prettylibrary}{\textcolor{library_color}{\textbf{Library}}}

\title{VISTA: Scale-Aware Visual Navigation via Action History Conditioning}

\author{
  Maeva Guerrier\\
  Polytechnique Montreal, MILA \\
  *equal contributions \\
  \texttt{maeva.guerrier@polymtl.ca} \\
  \And
  Koki Kobayashi \\
  Institute of Science Tokyo \\
  *equal contributions  \\
  \texttt{kobayashi.k.f785@m.isct.ac.jp}
  \And
  Simon Roy \\
  Polytechnique Montreal \\
  \And
  Jana Pavlasek \\
  Polytechnique Montreal, CoRA Lab \\
  \And
  Giovanni Beltrame \\
  Polytechnique Montreal, Mist Lab \\
}

\begin{document}


\maketitle



\begin{abstract}
    Vision Navigation Foundation Models (VNMs) promise end-to-end learned navigation policies capable of zero-shot deployment across diverse embodiments and environments. To maintain generality, many vision-based navigation models predict normalized actions. However, this normalization introduces a critical deployment vulnerability: applying different scaling factors to the same normalized trajectory alters its physical geometry, which degrades navigation performance and increases collision risks. We address this vulnerability by conditioning the model on normalized action histories alongside image observations, providing explicit context on the relationship between the model's predictions and the robot's actual physical displacement. Furthermore, current VNMs often struggle in visually repetitive environments that lack distinct features. To resolve this issue, we integrate a DINOv3 encoder, whose richer representations enable our model to capture both spatial and geometric dimensions between observations. VISTA generalizes robustly to out-of-distribution environments, achieving 100\% goal prediction accuracy in zero-shot, real-world deployment in Outdoor, Forest and Office settings, and an average of 95\% checkpoints crossed, demonstrating consistent path following in unseen environments.  
\end{abstract}



\section{Introduction}
Foundation models have become cornerstones of general-purposes policies that can be deployed zero-shot across diverse robot embodiments (e.g. Different cameras, sensors and robot types), novel environments and settings (e.g. navigation tasks, pick and place). Trained at large scale on large robotic datasets~\citep{walke2024bridgedatav2datasetrobot, khazatsky2025droidlargescaleinthewildrobot}, such models produce generalizable policies that transfer across settings without environment-specific calibration, a capacity classical hand-designed controllers fundamentally lack, as they cannot scale to the combinatorial diversity of real-world deployment conditions. 

Recent work, including ViNT~\citep{vint}, NoMaD~\citep{nomad}, MetricNet~\citep{nayak2025metricnet}, have demonstrated promising zero-shot visual navigation~\citep{guerrier2026visionfoundationmodelsnavigate} across embodiments. Vision foundation models enables the policy to learn prior understanding of diverse environments through large scale datasets and map them to waypoints for navigation. However, current approaches~\citep{gnm, vint, nomad} achieves embodiment generalization by normalizing the trajectories, which creates ambiguity at deployment. To maintain embodiment generality, vision-based navigation models predict normalized actions, where 2D waypoints are divided by the average training dataset waypoint spacing. However, without explicit knowledge of the intended physical step size, the model must infer it solely from sequential image observations. 

We demonstrate that conventional models fail at correctly estimating the scale of their predictions, and that visual inputs alone do not provide enough information to the model for it to adjust its predictions to match the correct trajectory. This ambiguity in scale inference leads to incorrect physical trajectories, degrading navigation performance.
To address this, we propose \textbf{VISTA}, Scale-Aware \textbf{Vis}ual Naviga\textbf{t}ion via \textbf{A}ction History Conditioning, a novel architecture and training objective that grounds action predictions in the physical scale by conditioning the policy on normalized action history. This allows VISTA to relate past normalized actions to observed visual displacement, generating metric-aligned waypoints robust to variations in robot speed and control frequency. 

\begin{figure*}[t]
\centering
\includegraphics[width=\linewidth]{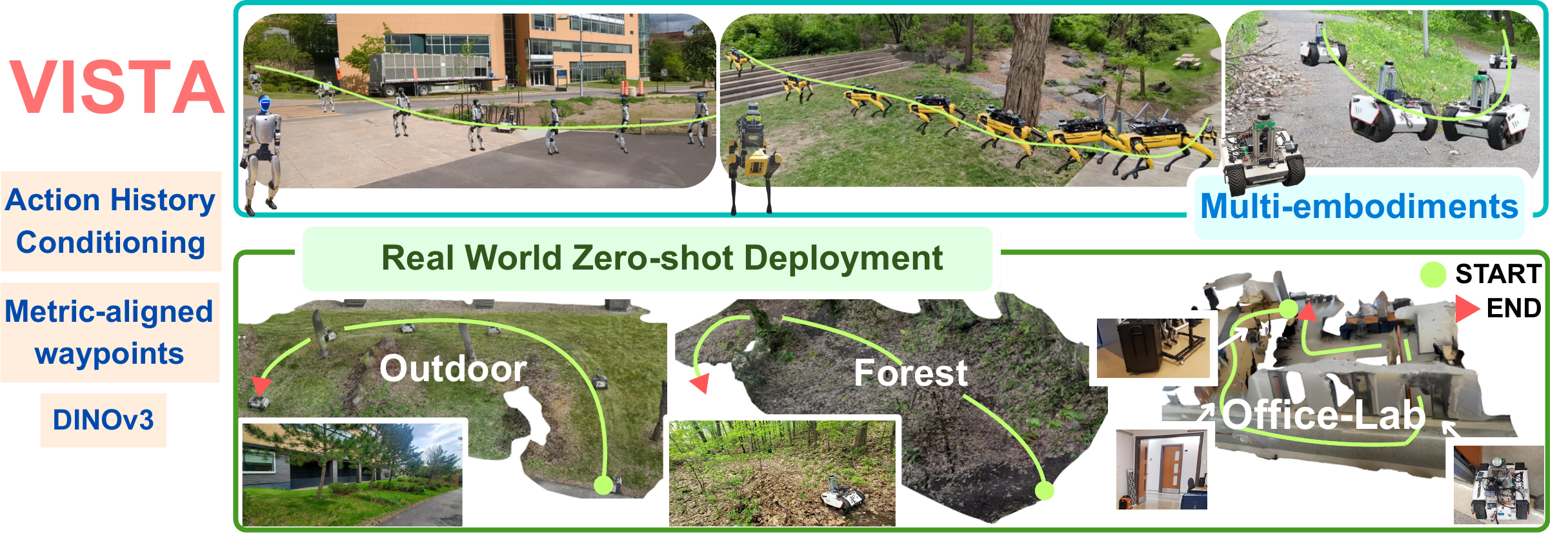}
\caption{Deployment time-lapse of VISTA: Scale-Aware Visual Navigation via Action History Conditioning, achieves zero-shot deployment in unseen environments by conditioning on action history to improve scaling across robotic platforms and trajectories.}
\label{fig:pitch}
\end{figure*}

Our key contributions include an analysis of metric misalignment in normalized visual navigation models, and VISTA (see Figure~\ref{fig:pitch}), a novel architecture conditioning on action history for metric-aligned waypoint prediction without scarifying cross-embodiment generality. We evaluate VISTA in zero-shot deployment across five real-world environments, demonstrating robustness to unseen deployment conditions.

We demonstrate that VISTA yields 100\% successful goal prediction in open outdoor space, office and forest settings and show that VISTA remains robust under perturbed conditions. We will make our code-base publicly available.


\section{Related Works}

A growing trend in robot navigation is the pursuit of zero-shot generalization~\citep{opennavQiao, zeroshot_objNav, Yin_2025_CVPR} across environments and embodiments~\citep{nomad, coal, navworldmodel, NavFormer}. Achieving this without environment-specific calibration requires learning based approaches~\citep{gnm, octo_2023, bharadhwaj2023roboagentgeneralizationefficiencyrobot}. GNM~\citep{gnm} pioneered large-scale visual navigation across diverse embodiments though an EfficientNet~\citep{Tan2019EfficientNetRM} encoder and feedforward architecture. Building on this, ViNT~\citep{vint} adopted a transformer backbone to capture long-horizon context and broaden generalization. However, vision-based navigation models predict normalized actions implicitly tied to average metric waypoint spacing seen during training, information unavailable at deployment, causing metric misalignment under speed or control frequency variations, a gap VISTA addresses through normalized action history conditioning. A recognized limitation~\citep{care, guerrier2026visionfoundationmodelsnavigate, zeng2025navidiffusorcostguideddiffusionmodel} of purely visual navigation models is their lack of explicit geometric understanding, cost-based approaches such as CARE~\citep{care} have emerged to address this, correcting trajectories online~\citep{zeng2025navidiffusorcostguideddiffusionmodel, cai2025navdplearningsimtorealnavigation}. A further limitation is encoder brittleness in featureless environments~\citep{guerrier2026visionfoundationmodelsnavigate} such as snow covered terrain or uniformed hallways, motivating the need of semantically rich visual encoders. DINOv3~\citep{dinov3} addresses this though self-supervised vision transformer trained on curated large-scale data. Precisely, its training strategy optimized high-level semantic understanding and fine-grained spatial features, yielding representations well-suited for navigation. VISTA builds on this foundation, adopting DINOv3 as its visual backbone.


\section{Problem Formulation}
Visual Navigation Models (VNMs) are navigation policies that output 2D waypoints 
$\hat{A}_t \in \mathbb{R}^{\ell \times 2}$ and a scalar distance prediction $\hat d_t \in \mathbb{R}$ 
from a sequence of $k$ observation images $O_t = \{o_{t-k+1}, \dots, o_t\}$ and a goal 
image $o_g$:
\begin{equation}
    f(O_t,\, o_g) \;\to\; (\hat{A}_t,\, \hat d_t)
\end{equation}

\textbf{Training:} VNM training datasets consist of paired observations and odometry 
poses $\{(o_i, p_i)\}_{i=1}^{T}$. Let $W_t := \{w_t, \dots, w_{t+\ell-1}\}$ be the future metric waypoints in the robot's local frame with prediction horizon $\ell$, where
\begin{equation}
    w_i = p_{i+1} - p_t, \qquad (t \leq i \leq t+\ell-1)
\end{equation}
The training target normalizes this sequence by the average inter-waypoint spacing 
$\phi_{\mathrm{train}}$ (the \emph{step size}):
\begin{equation}
    A_t := W_t \;/\; \phi_{\mathrm{train}}
\end{equation}

\textbf{Deployment:} When the goal image is outside the field of view, VNMs rely on a 
topological map $\mathbf{M} = (\mathcal{V}, \mathcal{E})$, where nodes $o_i \in \mathcal{V}$ 
are pre-collected observations and edges $e_{i,j} \in \mathcal{E}$ represent traversable 
connections. A belief over the closest node is maintained using the predicted temporal 
distance $\hat d_t$ to select a subgoal image $o_g$. Since predictions are normalized, 
waypoints are unnormalized at deployment as
\begin{equation}
    \hat{W}_t = \gamma_{\mathrm{deploy}} \times \hat{A}_t, 
    \qquad \gamma_{\mathrm{deploy}} = v_{\max} / f_{\mathrm{ctrl}}
\end{equation}
where $v_{\max}$ (m/s) is the maximum linear speed and $f_{\mathrm{ctrl}}$ (Hz) is the 
robot control frequency.

\textbf{Action Denormalization}: Since the real world geometry of the recovered metric trajectory will depend entirely on the deployment step size, the underlying normalized trajectory must be consistent with the chosen step size for the resulting waypoints to be metric-aligned. However, conventional VNMs are only conditioned on visual observations and a goal image, which does not provide enough information for them to infer step sizes reliably. 



\section{VISTA Architecture}
Our method modifies conventional VNM models mainly in two ways, namely the use of DINOv3~\citep{dinov3} and action history conditioning.

\textbf{DINOv3 Tokenization.} We use DINOv3 ViT-S, whose self-supervised representation captures geometric and semantic structure needed to understand inter-frame displacement and observation-to-goal relationship. 
Sequences of observation images and goal image are processed by DINOv3 to be converted into patch tokens.

Patch tokens are then projected 2x down-sampled to align feature dimensions while reducing computation. Context length and prediction horizon follow prior work~\citep{vint}.
 
\textbf{Action-history Conditioning.} We define the normalized action history as 
$H_t = \{a_{t-k+1}, \ldots, a_{t-1}\}$, where each $a_i$ is a past robot displacement normalized by the step size (see App.~\ref{sec:supp_action_history} for details). An MLP encodes the action history into tokens of the same dimension. By conditioning on the normalized action history, the model can learn to relate the observed visual displacement to physical step size, enabling metric-aligned waypoints predictions robust to variations in robot speed and control frequency, as demonstrated in Table~\ref{tab:varied_stepsize_results}.

\textbf{Transformer Decoder.} The processed tokens are concatenated with readout tokens for distance and action prediction, then passed through the transformer. Learnable temporal embeddings are added to the tokens. 2D RoPE is applied to image patch tokens following~\citep{oquab2024dinov2learningrobustvisual}.

\subsection{Normalized Metric Training} 

\textbf{Waypoint Scaling Factors.} To further condition the model on specific step sizes, we compute the step size for a sample as the average waypoint spacing of its individual trajectory. \textbf{Loss Function.} We use a multi-task loss combining action and distance objectives, balanced by a loss-balancing coefficient $\lambda \in [0,1]$, with $\mu=0.01$ used to match their scales:
\(    L_{\text{total}} = \lambda L_{\text{action}} + \mu(1-\lambda)L_{\text{distance}}.\)
$L_{\text{action}}$ is the MSE between predicted and ground-truth normalized robot-centered waypoints over $\ell$ step prediction horizon: 
\begin{equation}
    L_{\text{action}} = \frac{1}{N_\ell} \sum_{i=1}^N \sum_{j=1}^{\ell} (w_{ij}/ {\phi_i} - \hat{a}_{ij})^2
\end{equation}
While prior work defines distance loss as MSE on \enquote{temporal distance}~\citep{gnm, vint,nomad}, we instead supervise the model to produce normalized metric distances between the observation and the goal:
\begin{equation}
L_{\text{distance}} = \frac{1}{N} \sum_{i=1}^N (\hat{d}_i - \frac{\lVert p_{t_i} - p_{g_i}\rVert}{\phi_i})^2 
\end{equation}
where $\hat{d}_i$ is the predicted normalized goal distance. $\lambda = 0.5$ was used following~\citep{vint}.

\textbf{Varied Spacing Training.} While conventional models are only trained on 4Hz and 3Hz, we vary the sampling frequency of datasets from 2 to 12Hz to train robustness to different speeds and control frequencies.

\begin{figure}[t]
    \centering

    \begin{subfigure}[t]{0.24\linewidth}
        \centering
        \includegraphics[width=\linewidth]{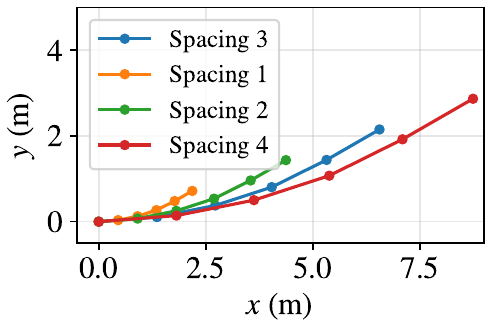}
        \caption{ViNT.}
        \label{fig:first}
    \end{subfigure}
    \hfill
    \begin{subfigure}[t]{0.24\linewidth}
        \centering
        \includegraphics[width=\linewidth]{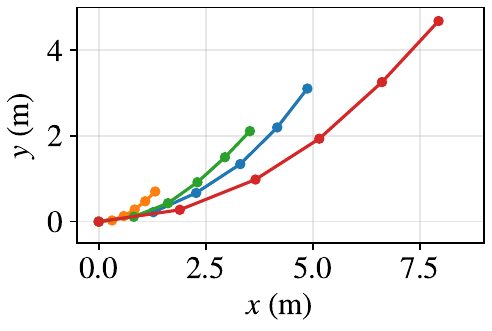}
        \caption{NoMaD.}
        \label{fig:second}
    \end{subfigure}
    \hfill
    \begin{subfigure}[t]{0.24\linewidth}
        \centering
        \includegraphics[width=\linewidth]{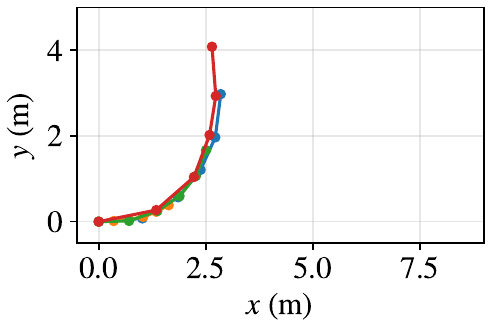}
        \caption{VISTA w/o AH.}
        \label{fig:third}
    \end{subfigure}
    \hfill
    \begin{subfigure}[t]{0.24\linewidth}
        \centering
        \includegraphics[width=\linewidth]{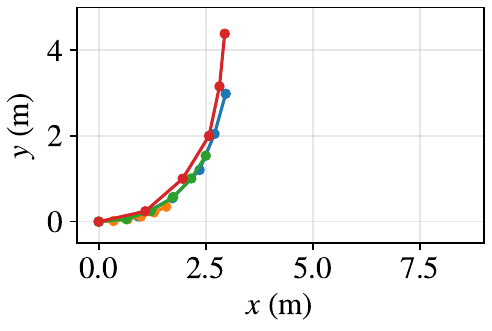}
        \caption{VISTA.}
        \label{fig:fourth}
    \end{subfigure}

    \caption{\textbf{Qualitative comparison of predictions on the same inputs.} The shape of the unnormalized trajectory stays the same regardless of observation spacing in ViNT. Due to the diffusion head, NoMaD produces trajectories with different shapes per spacing, but they fail to have the same curvature in metric space. VISTA models show much better alignment in metric space.}
    \label{fig:side_by_side}
\end{figure}

\begin{table}[t]
\centering
\small
\setlength{\tabcolsep}{4pt}
\begin{tabular}{l rr rrr rrr rrr}
\toprule
& \multicolumn{2}{c}{Prediction Losses}
& \multicolumn{6}{c}{Cross-Track Error by Obs.\ Step size} \\
\cmidrule(lr){2-3}
\cmidrule(lr){4-9}
& & & \multicolumn{2}{c}{Step size 1}
    & \multicolumn{2}{c}{Step size 2}
    & \multicolumn{2}{c}{Step size 4} \\
\cmidrule(lr){4-5}
\cmidrule(lr){6-7}
\cmidrule(lr){8-9}
Method & Dist. & Act. & Avg & Worst & Avg & Worst & Avg & Worst \\
\midrule
ViNT          & 11.581 & 0.589 & 0.020 & 0.054 & 0.019 & 0.050 & 0.021 & 0.054 \\
NoMaD          & 22.380 & 0.879 & 0.024 & 0.063 & 0.042 & 0.105 & 0.066 & 0.162 \\
VISTA          &  \textbf{3.595} & \textbf{0.217} & 0.011 & 0.027 & \textbf{0.011} & 0.028 & 0.016 & 0.038 \\
VISTA w/o AH        &  6.786 & 0.293 & \textbf{0.009} & \textbf{0.023} & \textbf{0.011} & \textbf{0.027} & \textbf{0.015} & \textbf{0.037} \\
\bottomrule
\end{tabular}
\vspace{2mm}
\caption{\textbf{Offline evaluation of prediction accuracy and scale consistency.} Cross-track error is computed between metric trajectories predicted under different observation step size, using observation step size 3 as reference. Prediction losses shows the loss of the prediction with spacing 3.}
\label{tab:cross_track}
\end{table}



\section{Results}
\subsection{Offline Evaluation of Scale Consistency}

\textbf{Experiment Setup:} We evaluate scale consistency by testing each model on the same current observation and goal image while varying the temporal spacing of the past observation sequence. We use the 12Hz version of SACSoN as the dataset, thus an observation spacing of 3 corresponds to the standard 4Hz setting. Further details are provided in App.~\ref{sec:supp_scale_consistency}.

\textbf{Evaluation Metric:} A model is scale-consistent if trajectories predicted under different observation spacings agree after being converted back to metric space. We compare each spacing with the standard spacing 3, computing cross-track error from the shorter metric trajectory to the longer one ($1,2 \!\rightarrow\! 3$ and $3 \!\rightarrow\! 4$). We report the mean and maximum waypoint-to-trajectory distances.

\textbf{Results:} Figure~\ref{fig:side_by_side} provides a qualitative comparison of the predictions, and Table~\ref{tab:cross_track} summarizes the offline evaluation. VISTA, and VISTA w/o AH reduce cross-track error compared to baselines across observation spacings, indicating improved consistency under changes in temporal spacing. Although VISTA w/o AH achieves better scale consistency, the full model obtains the lowest prediction losses, suggesting that action history improves trajectory prediction accuracy while preserving the scale consistency obtained from the stronger visual encoder.

\begin{figure}[t]
    \centering
    \includegraphics[width=0.85\linewidth]{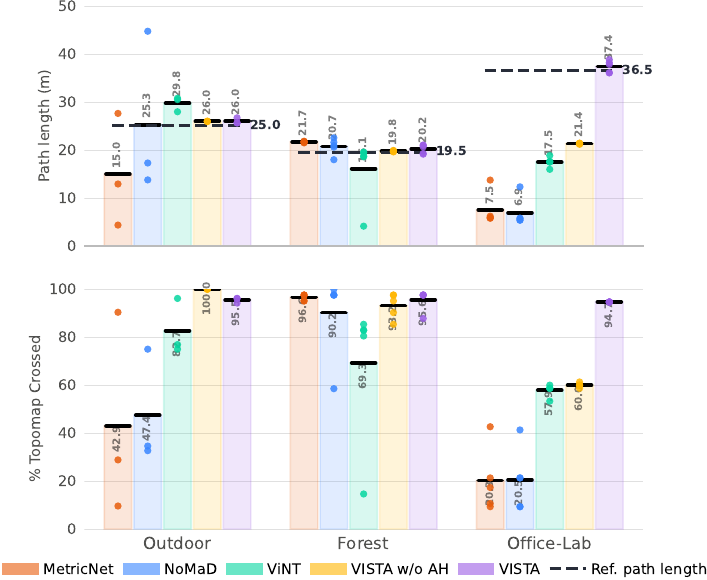}
    \caption{\textbf{Quantitative results:} Path length and topological map (topomap) crossed, across scenes.}
    \label{fig:real_world_pathlen_navpercent}
\end{figure}

\begin{table}[t]
\centering
\small
\caption{\textbf{Performance in three environments.} Outcomes are classified as: \textit{Success} (green: terminal position error $\leq$ 2 meters), \textit{Incomplete} (gray: false-positive goal reach prediction), \textit{Collision} (red: a collision occurred). Results denote $n_{\text{outcome}} / \text{number of trials}$.}
\label{tab:outcomes_sr_coll_inc_outdoor_forest_lab}
\begin{tabular}{@{}l c c c c c @{}}
\toprule
\textbf{Environments} & \textbf{VISTA} & VISTA w/o AH & ViNT & NoMaD & MetricNet \\ 
\midrule
\textbf{Outdoor} & \fullcell{mygreen}{3/3} & \fullcell{mygreen}{3/3} & \splitcellthree{mygreen}{0.33}{1/3}{myred}{0.66}{1/3}{mygrey}{0}{1/3} & \splitcelltwo{myred}{0.67}{2/3}{mygrey}{0.33}{1/3} & \splitcelltwo{myred}{0.33}{1/3}{mygrey}{0.67}{2/3} \\ \addlinespace
\textbf{Forest}  & \fullcell{mygreen}{5/5} & \splitcelltwo{mygreen}{0.60}{3/5}{mygrey}{0.40}{2/5} & \splitcelltwo{mygrey}{0.80}{4/5}{myred}{0.20}{1/5} & \splitcelltwo{mygreen}{0.80}{4/5}{myred}{0.20}{1/5} & \splitcelltwo{mygreen}{0.60}{3/5}{mygrey}{0.40}{2/5} \\ \addlinespace
\textbf{Office-Lab}  & \fullcell{mygreen}{5/5} & \fullcell{myred}{5/5} & \fullcell{myred}{5/5} & \fullcell{myred}{5/5} & \fullcell{myred}{5/5} \\ 
\bottomrule
\end{tabular}
\end{table}

\subsection{Real-World Evaluation Setup}

The metrics include collision occurrence, successes, incompleteness, path length and percentage of topological map navigated (PTN). A trial is successful if the robot reaches within 2 meters (m) of the goal. Trials that neither succeed nor collide are marked incomplete. Collisions are terminal events, bounding each trial to at most one. Path length is the total distance traveled from start to termination. For PTN, we place checkpoints every 0.5 m along the reference trajectory. A checkpoint is considered passed if the robot is within 2 m of it.
The deployment environment are: \textit{Outdoor} (25 m path): Grass field bordered by symmetric rows of bushes (see App.~\ref{sec:supp_environments}). The open layout minimizes collision risk, making this primarily a test of goal localization and path following. \textit{Forest} (19 m path): visually dense setting with foliage, logs, and trees. Straying away from the path risks collision. \textit{Office-Lab} (36 m path): The path passes through an open workshop, then a corridor of visually identical doors, targeting the third, and ends with two tight doorframes requiring strict path following. As for the baselines, all methods, including VISTA, are deployed zero-shot without fine-tuning. We compare against ViNT~\citep{vint}, NoMaD~\citep{nomad}, MetricNet~\citep{nayak2025metricnet} and VISTA without action history (VISTA w/o AH). We include MetricNet but exclude MetricNav, whose cost-based diffusion objective is specially designed for obstacle avoidance, an advantage not shared by any other baselines, including ours. As no deployment code was publicly available for MetricNet at the time of our experiments, we followed the author's implementation guideline.

\begin{wrapfigure}{r}{0.31\textwidth}
    \centering
    \includegraphics[width=\linewidth]{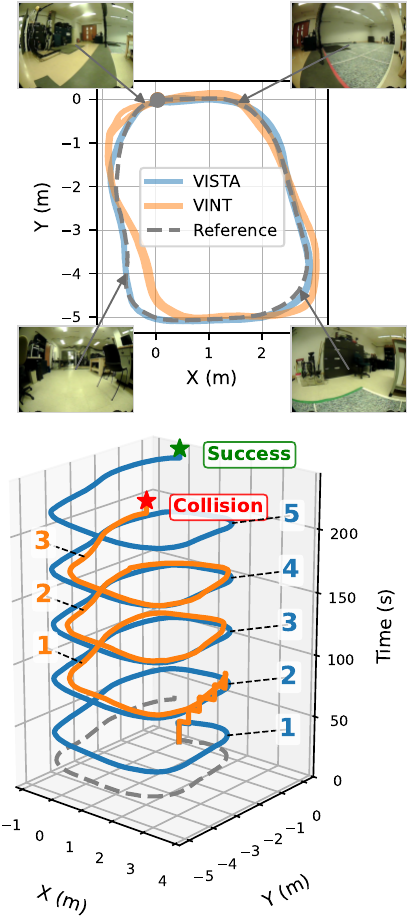}
    \caption{\textbf{Visual of the loop experiments.} VISTA reliably achieves 5 loops across 5 trials.}
    \label{fig:loop_results_fig}
\end{wrapfigure}

\subsection{Deployment under varied step sizes}

We evaluate sharp-turn execution in a hallway while varying the robot’s effective step size to see its robustness. Specifically, we change the control frequency $f$ and maximum linear velocity $v_\text{max}$, since $v_\text{max} / f$ determines how far the robot moves between each timestep. We test control frequencies of 4, 2, and 1 Hz and maximum linear velocities of 0.1, 0.2, and 0.4 m/s, covering effective step sizes from 0.4m to 0.025m.

Table~\ref{tab:varied_stepsize_results} reports successes and collisions under varied step sizes. VISTA successfully completes the maneuver in most cases, with degradation at the extreme 1 Hz setting and a single collision in 0.2 m/s trials, which occurred after successfully executing the sharp turn. ViNT succeeds in default settings but consistently fails at low speed, colliding before completing the turn, showing it struggles to adapt to step size changes. Action History Conditioning gives VISTA a reference to the step size, enabling waypoint prediction that remains physically consistent across setups. 

\subsection{Zero-shot deployment}

Table~\ref{tab:outcomes_sr_coll_inc_outdoor_forest_lab} reports the deployment trial outcomes, namely success, incompleteness and collisions, while Figure~\ref{fig:real_world_pathlen_navpercent} shows the average path length of each of the baselines with respect to the reference path length and reports the percentage of topological map navigated.

\textbf{Outdoor.} VISTA and VISTA w/o AH achieve 95\% and 100\% PTN (see Outdoor Figure~\ref{fig:real_world_pathlen_navpercent}), with consistent path length and accurate goal prediction across all trials. ViNT and NoMaD diverge from the reference trajectory, inaccurately predicting goal completion (see Outdoor in Table~\ref{tab:outcomes_sr_coll_inc_outdoor_forest_lab}), a localization failure we attribute to weaker visual encoders, which lack sufficient representational capacity in scenes with low distinctiveness. NoMaD collides in 2/3 trials; the remaining trial ends incompletely, exposing an inability to recover from accumulated positional errors. MetricNet, inheriting NoMaD's encoder with rescaled metric outputs, reduces collisions (1/3 trials) but produces false goal prediction in 2/3 trials. Together, these results reveal clear separation: VISTA and VISTA w/o AH maintain reliable path following under perceptual aliasing, while ViNT, NoMaD, and MetricNet exhibit trajectory drift, false goal prediction, and collisions.

\WFclear 
\textbf{Forest.} In this feature-rich environment, VISTA, VISTA w/o AH, NoMaD and MetricNet remain close to the reference path length (see Forest in Figure~\ref{fig:real_world_pathlen_navpercent}). Yet, only VISTA achieves 100\% accurate goal prediction (see Table ~\ref{tab:outcomes_sr_coll_inc_outdoor_forest_lab}), suggesting that richer visual features benefit trajectory following but do not resolve goal prediction failures in weaker encoders. 

\textbf{Office-Lab.} VISTA achieves 100\% success (5/5 trials) (see Office-Lab in Table~\ref{tab:outcomes_sr_coll_inc_outdoor_forest_lab}), reliably clearing tight doorways and correctly identifying the third corridor door. VISTA w/o AH correctly identifies the target door but collides systematically across all trials. We attribute this failure to the absence of AH, as waypoint generation becomes sensitive to velocity variations, which is critical when decelerating through tight turns. ViNT targets the second door rather than the third in 4/5 trials, colliding with the wall, whereas in the remaining trial, ViNT identifies the correct door but still ends in collision. NoMaD and MetricNet fail globally (0/5 trials), colliding with objects in the workshop upon entry.

These results demonstrate that encoders with generalized representational capacities, such as DINOv3 are necessary for reliable visual navigation. However, representation quality alone is insufficient; metric-scale motion context provided through action history is additionally required for reliable end-to-end navigation.


\section{Stress-testing VISTA: Perturbation Robustness}
\begin{figure}[b]
    \centering
    \includegraphics[width=\linewidth]{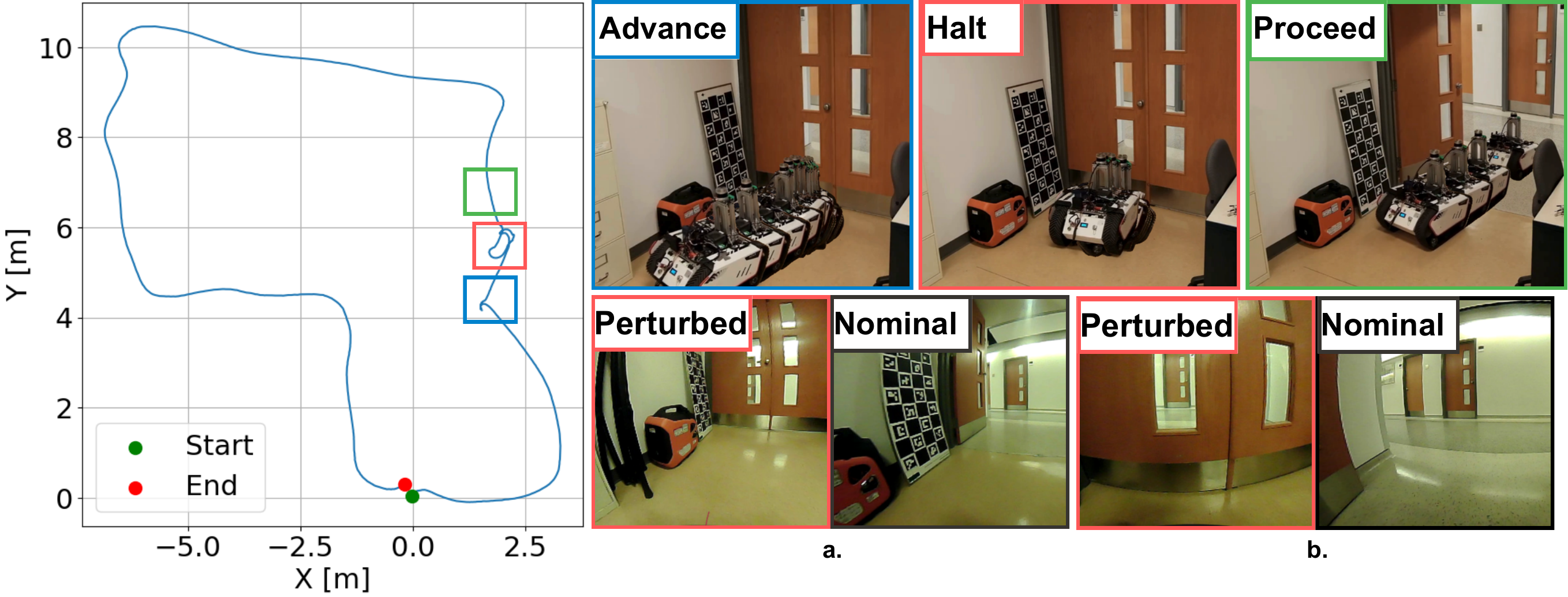}
    \caption{\textbf{Dynamic perturbation.} Our model navigates the Office-Lab environment using a topological map built under nominal conditions (\textit{door open}). At deployment time, the door is closed, then reopened. Across all \textit{5 trials}, this halt is observed.}
    \label{fig:fig_openclose_explained}
\end{figure}

\textbf{Dynamic perturbation:} We tested this by deploying VISTA with a topological map showing an open door, closed at deployment (see Figure~\ref{fig:fig_openclose_explained}). Across 5 trials VISTA decelerated and halted the door was open again(see~\ref{sec:supp_environments}, we note that 1 trial had a failure which an unrelated collision later in the path), demonstrating robustness against topological map inconsistency.

\begin{table}[t]
\centering
\small
\caption{\textbf{Robustness to deployment step size.}
We vary the effective step size (Step. S) by varying the control frequency or the maximum
linear velocity. (A): 0.4m/s / 4Hz, (B): (0.4m/s / 2Hz), (C): (0.4m/s / 1Hz), (D): (0.2m/s / 4Hz) , (E): (0.1m/s / 4hz)
MetricNet is excluded as it fails under the default configuration. Green indicates
successes and red indicates collisions.}
\label{tab:varied_stepsize_results}
\setlength{\tabcolsep}{4pt}
\begin{tabular}{l ccc p{0.3cm} cc}
\toprule
& \multicolumn{1}{c}{\textbf{Default}}
& \multicolumn{2}{c}{\textbf{Varied Ctrl. Freq.}}
&& \multicolumn{2}{c}{\textbf{Varied Max. Lin. Vel.}} \\
\cmidrule(r){2-2}
\cmidrule(r){3-4}
\cmidrule(l){6-7}
\diagbox[width=2.3cm,height=0.6cm]{\textbf{Method}}{\textbf{Step. S}}
& \textbf{\begin{tabular}{c}0.10m \\ (A)\end{tabular}}
& \textbf{\begin{tabular}{c}0.20m \\ (B)\end{tabular}}
& \textbf{\begin{tabular}{c}0.40m \\ (C)\end{tabular}}
&&
\textbf{\begin{tabular}{c}0.05m \\ (D)\end{tabular}}
& \textbf{\begin{tabular}{c}0.025m \\ (E)\end{tabular}} \\
\midrule
\textbf{VISTA}
& \fullcell{mygreen}{3/3}
& \fullcell{mygreen}{3/3}
& \fullcell{myred}{0/3}
&&
\splitcelltwo{myred}{0.33}{1/3}{mygreen}{0.67}{2/3}
& \fullcell{mygreen}{3/3} \\

ViNT
& \fullcell{mygreen}{3/3}
& \splitcelltwo{myred}{0.33}{1/3}{mygreen}{0.67}{2/3}
& \fullcell{myred}{0/3}
&&
\fullcell{myred}{0/3}
& \fullcell{myred}{0/3} \\

MetricNet
& \fullcell{myred}{0/3}
& --
& --
&&
--
& -- \\
\bottomrule
\end{tabular}
\end{table}

\textbf{Experimenting distance consistency:} In VNMs, subgoal/goal selection depends on predicted distances, inconsistent predictions cause subgoal mismatches that compound over time. We evaluate distance consistency via a repeated-loop experiment (see~\ref{sec:supp_loop}), a topological map is built over a single loop and 5 consecutive loops must be completed. VISTA succeeds in all 5 trials. 

\textbf{Beyond RGB images, the strength of DINOv3:} Table~\ref{tab:depth_results} reports losses on depth images. VISTA achieves the lowest loss across all loss metrics, showcasing its capability to handle depth inputs. We provide more details in \ref{sec:supp_depth}.

\begin{table}[t]
\centering
\small
\caption{\textbf{Report of losses on depth images.} Comparison of distance, action and total loss on unseen dataset (Office) with depth images as input.}
\label{tab:depth_results}
\begin{tabular}{lccc}
\toprule
\textbf{Baselines} & \textbf{Distance Loss} & \textbf{Action Loss} & \textbf{Total Loss} \\
\midrule
ViNT w/o depth & $ 21.860 \pm 11.461 $ & $ 0.328 \pm 0.203 $ & $ 0.273 \pm 0.148 $ \\
\textbf{VISTA} w/o depth & \textbf{12.733} $\pm$ \textbf{12.911} & \textbf{0.1828} $\pm$ \textbf{0.1013} & \textbf{0.1551} $\pm$ \textbf{0.0831}  \\
ViNT w/ depth  & $ 70.858 \pm 19.373 $ & $ 0.708 \pm 0.278 $ & $ 0.708 \pm 0.146 $ \\
\textbf{VISTA} w/ depth & \textbf{57.420} $\pm$ \textbf{23.632} & \textbf{0.282} $\pm$ \textbf{0.176} & \textbf{0.428} $\pm$ \textbf{0.166}  \\
\bottomrule
\end{tabular}
\end{table}

\section{Conclusion and Discussion}
We introduce VISTA, an approach that enables metric-aligned waypoints, grounded in physical scale via normalized action history, enabling robust navigation across varying control frequencies, and robot speed. VISTA achieves reliable  zero-shot deployment in unseen environments, with accurate goal prediction across five distinct real-world setting: open outdoor, forest, office, hallway and workshop. 

\section{Limitations}
Despite promising results, our approach has three main limitations. First, VISTA operates without explicit knowledge of the robot's physical dimensions or embodiment-specific parameters, which may limit control in scenarios where geometry is critical, such as assessing traversability. As an example, a model piloting a rover that has its embodiment-specific knowledge would avoid stairs while trying to align with the requested trajectory. This could be addressed by conditioning the model on a traversability map during learning.


Second, in visually similar settings where consecutive past observations are nearly identical, our model can struggle to infer the required motion from visual input alone. This motivates the future integration of adjustable topomap sampling to avoid images that are either too close together or too far apart, thus eliminating trajectory confusion. 


Third, at very low control frequencies ($\leq$ 1 Hz), the absence of explicit robot geometry makes it difficult to ground motion magnitude from observations alone. 



\clearpage



\clearpage
\section{Supplementary materials}


\subsection{Method Details}
\label{sec:supp_method}

\begin{figure}[b!]
    \centering
    \includegraphics[width=0.85\textwidth]{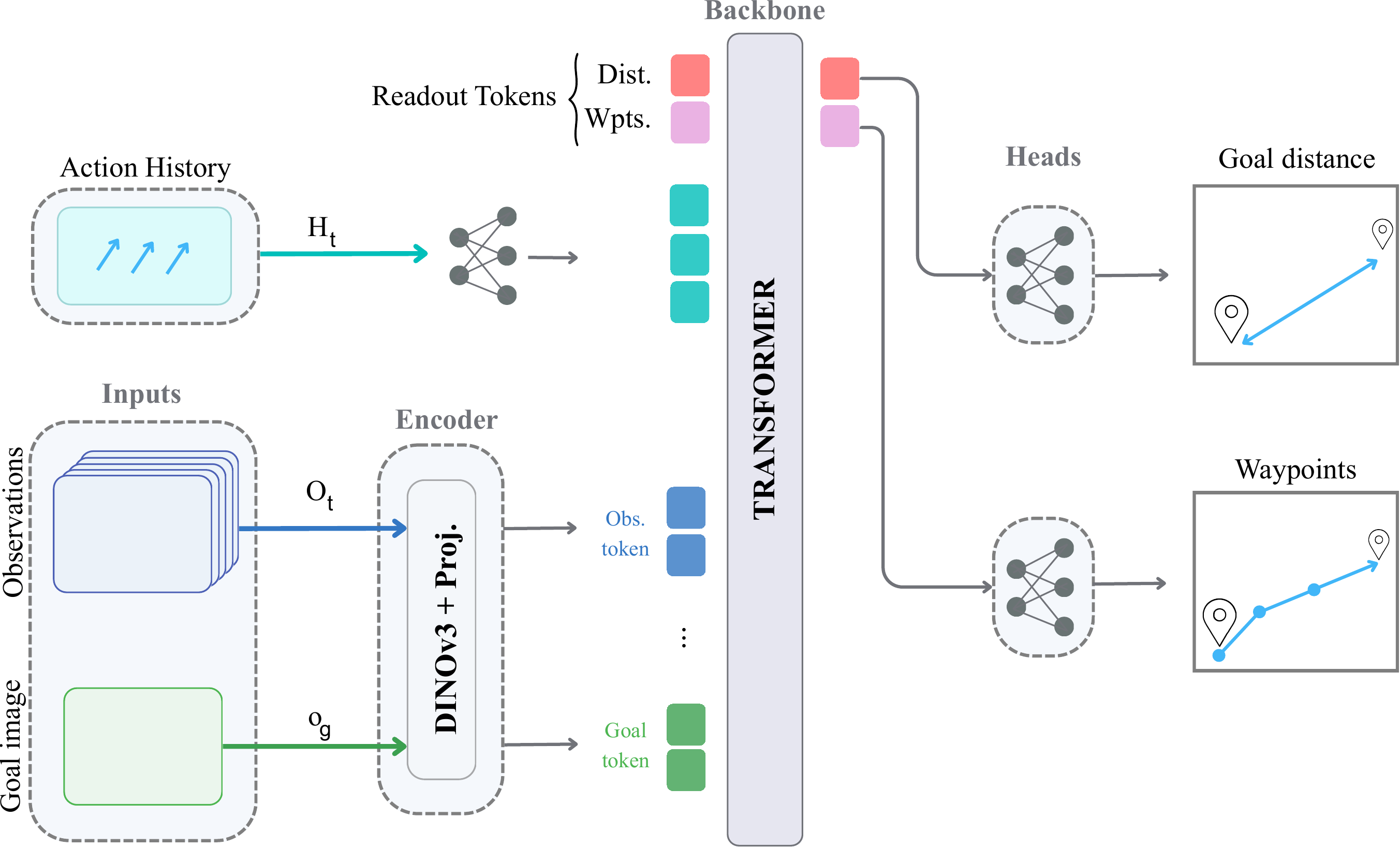}
    \caption{VISTA takes as input a sequence of image observations $O_t$ and a goal image $o_g$ at timestep $t$. The DINOv3 encoder encodes the observation sequence and the goal observation, respectively. The resulting patch tokens are then processed by the projector network, matching feature dimension while downsampling them by 2x. The backbone is composed of 6 transformer layers with 512 dimensions, 8 attention heads, and a feed-forward dimension of $4\times512$. Our proposed architecture outputs waypoints $\hat{w}_{1:T}$, alongside metric distance $\hat{d}$. }
    \label{fig:architecture_vista}
\end{figure}

\subsubsection{Goal-conditioned waypoint prediction and target normalization}
\label{sec:supp_normalization}

Throughout this section, index ranges are inclusive: $x_{i:j}$ denotes the sequence $(x_i, x_{i+1}, \ldots, x_j)$. Given a trajectory $\tau$ defined as a sequence of visual observations and poses over $T$ timesteps,
$
\tau := \{(o_i, p_i)\}_{i=1}^{T}.
$
Let $W_t := w_{t:t+\ell-1}$ denote the future metric waypoint sequence in the robot's local frame, where $\ell$ is the prediction horizon. Here, each $w_i$ is a single metric waypoint, while $W_t$ denotes the entire sequence of future waypoints starting from time $t$.

Conventional VNMs' actions predictions are sequences of 2D waypoints defined as$\hat{A}_t \leftarrow f_{\theta}(O_t,o_g)$ where $k$ is the observation context length, $o_g$ is the goal or subgoal RGB image, and $\hat{A}_t$ is the predicted normalized waypoint sequence. The training target is defined as $A_t := W_t / \phi_{\mathrm{train}},$ where $\phi_{\mathrm{train}}$ is the average waypoint spacing between consecutive waypoints in a trajectory, which we call \textit{step size}. It determines the physical scale of the prediction target: larger step sizes span more metric distance per prediction step, while smaller ones yield denser trajectories.

\subsubsection{Deployment-time denormalization}
\label{sec:supp_denorm}

During deployment, the model output must be converted back into physical, executable waypoints. Given the normalized prediction $\hat{A}_t$, the corresponding metric waypoint sequence is recovered using a deployment step size $\phi_{\mathrm{deploy}}$:
$\hat{W}_t = \phi_{\mathrm{deploy}} \times \hat{A}_t .$ Here, $\phi_{\text{deploy}}$ is a configurable parameter set during deployment, usually determined by the robot's maximum speed $v_\text{max}$ and control frequency $f$ as $v_\text{max} / f$. Since the step size governs both speed and curve geometry, path following requires predictions that explicitly account for it, yet conventional models lack such a reference, relying instead on the visual observation sequence $o_{t-k+1:t}$ alone to implicitly infer the required adjustments.

A naive solution assumes all displacements correspond to $\phi_\text{deploy}$, but this breaks when the robot deviates from its standard speed. A model relying solely on visual input cannot distinguish a fast robot temporarily slowing down from a slow robot at full speed. If distinct physical states produce the same visual change, the model cannot reliably estimate $\phi_\mathrm{deploy}$ and align its prediction with the real world. To resolve this without sacrificing the embodiment generality of normalized action, we introduce the concept of action history conditioning.

\subsubsection{Action history contextualization}
\label{sec:supp_action_history}

Let $a_{t-k+1:t-1}$ be the past $k-1$ actions, where an individual past action is defined as $a_i = \frac{\Delta p_i}{\phi_{\text{deploy}}}$. Our model $f_{\text{vista}}$ is formulated as: $f_{\text{vista}}(o_{t-k+1:t}, a_{t-k+1:t-1}, o_g) \to \frac{w_{t:t+\ell-1}}{\phi_{\text{deploy}}}$. By conditioning on the normalized action history $a_{t-k+1:t-1}$, the model learns to relate the observed visual displacement to physical step size, enabling metric-aligned waypoints predictions robust to variations in robot speed and control frequency.

\subsubsection{Design Choices}
\label{sec:supp_design}

A key design choice in our method is to keep both the conditioning signal and the prediction target in the normalized action space. One possible approach to the scale ambiguity problem is to condition the model explicitly on the step size. However, this requires the model to learn how a metric scalar should affect a trajectory represented in normalized, robot-centered coordinates. In practice, this introduces a difficult cross-space grounding problem between metric scale and normalized action prediction.

Instead, we condition the model on recent normalized action history. This provides a reference for the robot's recent motion scale in the same representation space as the output waypoints. As a result, the model can adapt its predicted curvature and displacement without explicitly translating from metric space to normalized action space.

Another advantage of supplying the model with action history is that the action history can serve as a baseline for interpreting visual displacement. This is analogous to multi-view relative depth estimation, where image correspondences alone determine structure only up to scale, and the camera baseline provides the reference needed to relate observed parallax to physical displacement. In our setting, the observation sequence contains visual changes caused by the robot’s recent motion, but those changes are ambiguous without knowing the scale of that motion. By providing recent normalized actions, the model receives an explicit motion baseline in the same coordinate system as the predicted waypoints, making it easier to interpret inter-frame displacement and infer scale-consistent future actions.

Finally, we use DINOv3 to strengthen the visual representation. While action history provides motion-scale context, the model still needs to reason about visual geometry, inter-frame displacement, and the relationship between the current and goal observations. DINOv3 provides stronger features for this visual reasoning. With these additions, the total number of parameters of the model is now 42M.

\subsection{Implementation Details}
\label{sec:supp_impl}

\subsubsection{Model and training parameters}
\label{sec:supp_training_params}

VISTA architecture is shown in Figure~\ref{fig:architecture_vista}. Table~\ref{tab:training_hyperparams} reports the training hyperparameters for VISTA, and Table~\ref{tab:model_config} reports the specific configuration regarding our architecture. 

\begin{table}[t!]
\centering
\small
\caption{Training hyperparameters.}
\label{tab:training_hyperparams}
\begin{tabular}{ll}
\toprule
Parameter & Value \\
\midrule
Batch size & 192 \\
Epochs & 10 \\
Optimizer & AdamW \\
Learning rate & $3.0 \times 10^{-4}$ \\
DINO learning-rate multiplier & 0.1 \\
Gradient clipping & True \\
Maximum gradient norm & 1.0 \\
Scheduler & Cosine Annealing \\
Warmup epochs & 3 \\
Minimum learning rate & $1.0 \times 10^{-6}$ \\
Warmup initial learning rate & $1.0 \times 10^{-5}$ \\
\bottomrule
\end{tabular}
\end{table}

\begin{table}[t!]
\centering
\small
\caption{Model configuration.}
\label{tab:model_config}
\begin{tabular}{ll}
\toprule
Parameter & Value \\
\midrule
Observation encoder & \texttt{facebook/dinov3-vits16-pretrain-lvd1689m} \\
Image size & $192 \times 192$ \\
Observation encoding size & 512 \\
Number of MHA heads & 8 \\
Number of MHA layers & 6 \\
MHA feed-forward dimension factor & 4 \\
Output layers & [256] \\
Use action history & True \\
Action encoder layers & [256] \\
Downsampling method & Pixel unshuffle \\
\bottomrule
\end{tabular}
\end{table}

Since we employed a normalized metric training objective for distance prediction, negative sampling~\citep{gnm} is no longer used during training.

\subsubsection{Deployment logic}
\label{sec:supp_deployment_logic}

We present an overview of the deployment pipeline in the pseudo-algorithm~\ref{alg:deploy_algo}.

\subsection{Experimental Setup}
\label{sec:supp_setup}

\subsubsection{Datasets}
\label{sec:supp_datasets}

\begin{figure}[b!]
    \centering
    \includegraphics[width=0.35\linewidth]{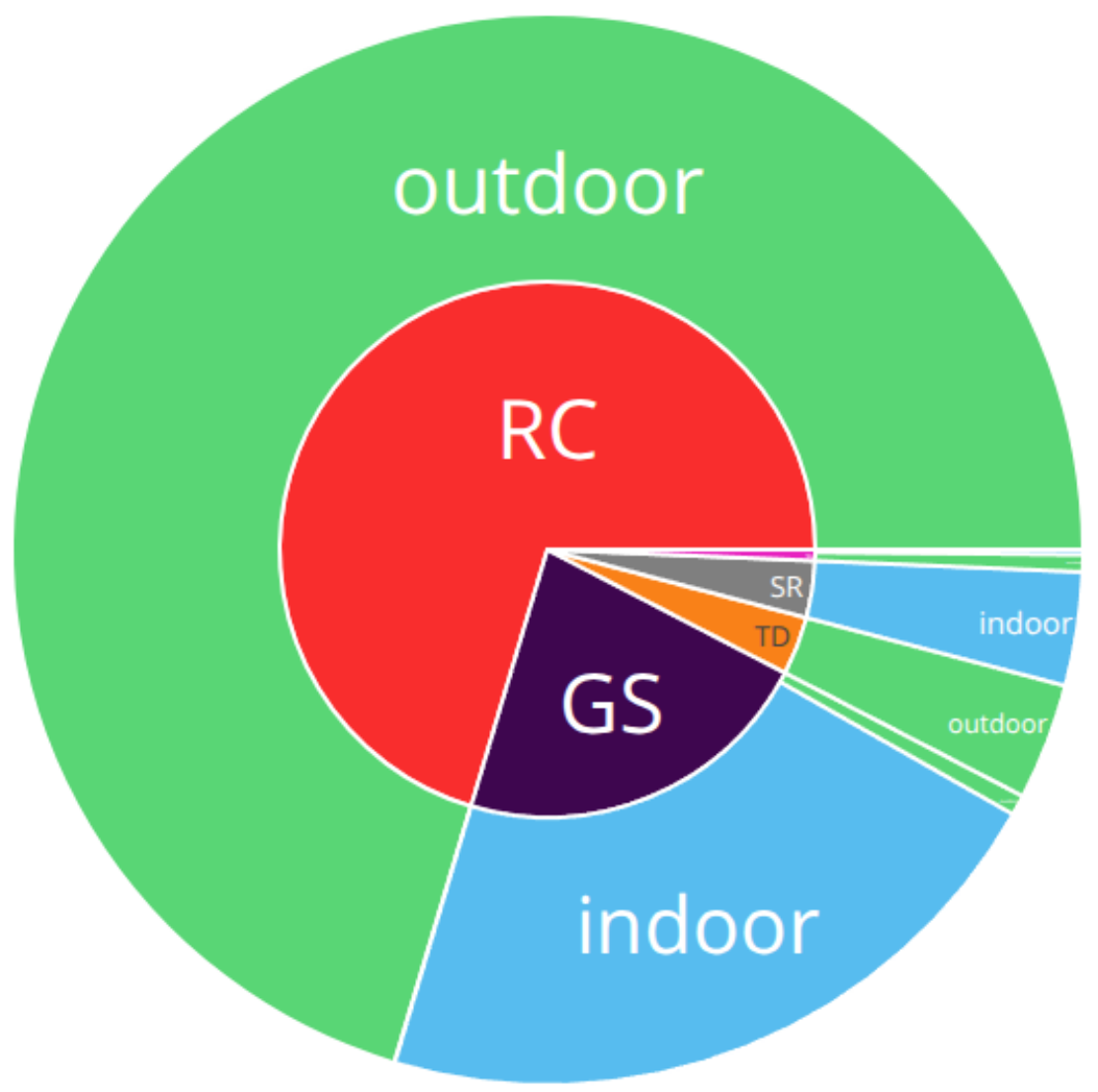}
    \includegraphics[width=0.35\linewidth]{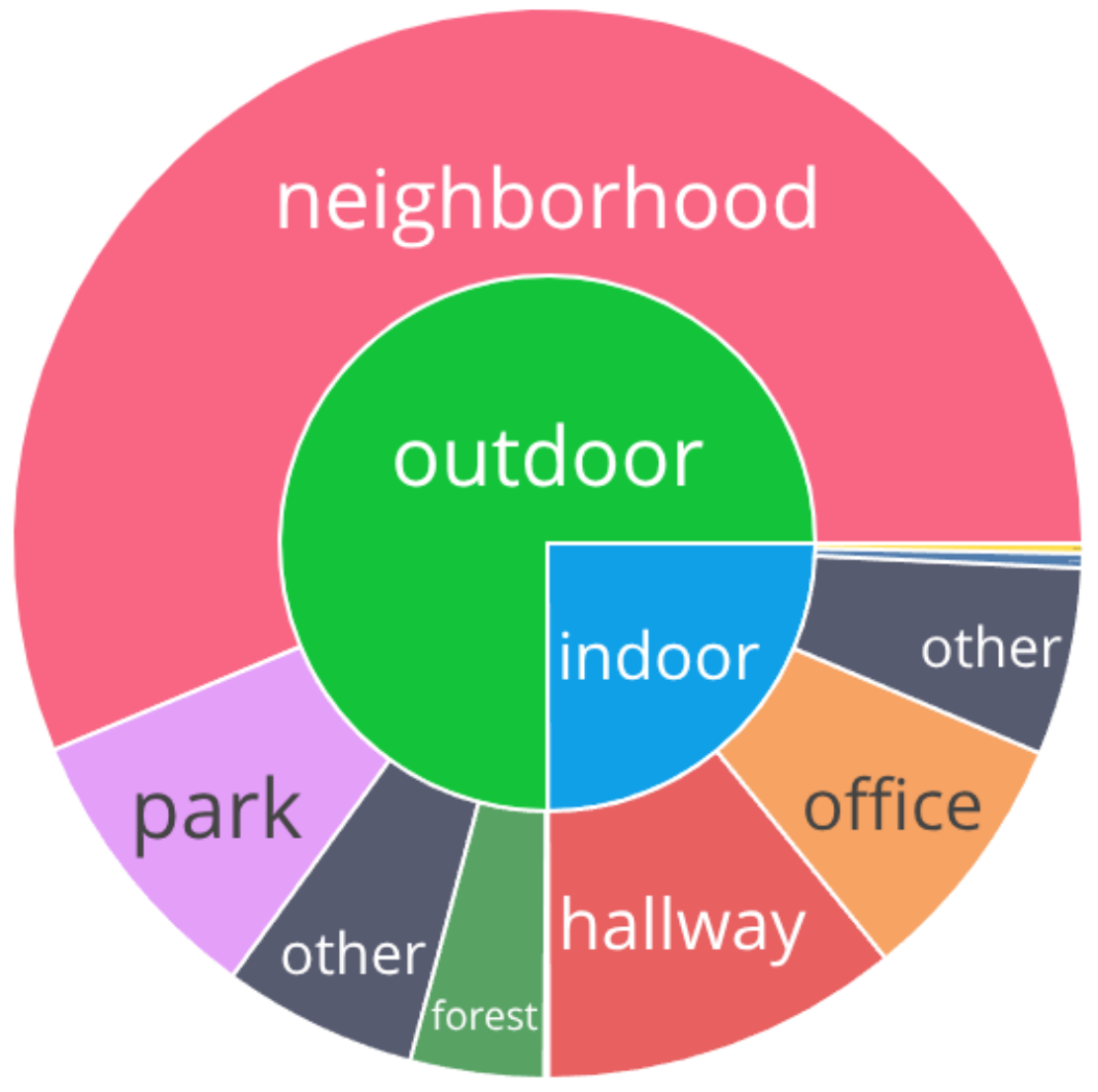}
    \caption{\textbf{Dataset environment composition.} Indoor-outdoor split per dataset (\textit{left}); \prettysacson{}, \prettyscand{}, \prettyrecon{}, \prettytartan{} and \prettystanford{}. Per-environment breakdown across indoor and outdoor settings (\textit{right}); where the two smaller slices are \prettygymnasium{} and \prettylibrary{}.}
    \label{fig:dataset_composition}
\end{figure}
For the training datasets, we used a combination of trajectory data obtained from GoStanford~\citep{go_stanford}, RECON~\citep{recon}, SACSoN~\citep{hirose2023sacson}, SCAND~\citep{karnan2022scand}, and TartanDrive~\citep{urmson2007tartan}. 

In order to get the datasets composition we used Qwen3-VL-4B-Instruct and the prompt reported in \textit{Qwen Input Prompt}.

\begin{promptbox}
Analyze these sequential images from a trajectory. 

Based on these keywords: \texttt{\{','.join(KEYWORDS)\}}, identify the environment type 
(e.g., office, forest) and list the objects present. 

Only use the keywords \texttt{\{', '.join(KEYWORDS)\}} and the following environment 
types: \texttt{\{', '.join(ENV\_TYPES)\}} to guide your analysis. 

Do not give any additional information that are not part of the keywords 
\texttt{\{', '.join(KEYWORDS)\}} or environment types \texttt{\{', '.join(ENV\_TYPES)\}}.

Output only a JSON-like format: 
\texttt{\{\{'env': 'type', 'objects': ['obj1', 'obj2'], 'location': 'indoor/outdoor'\}\}}
\end{promptbox}

\begin{figure}[b!]
    \centering
    \includegraphics[width=\linewidth]{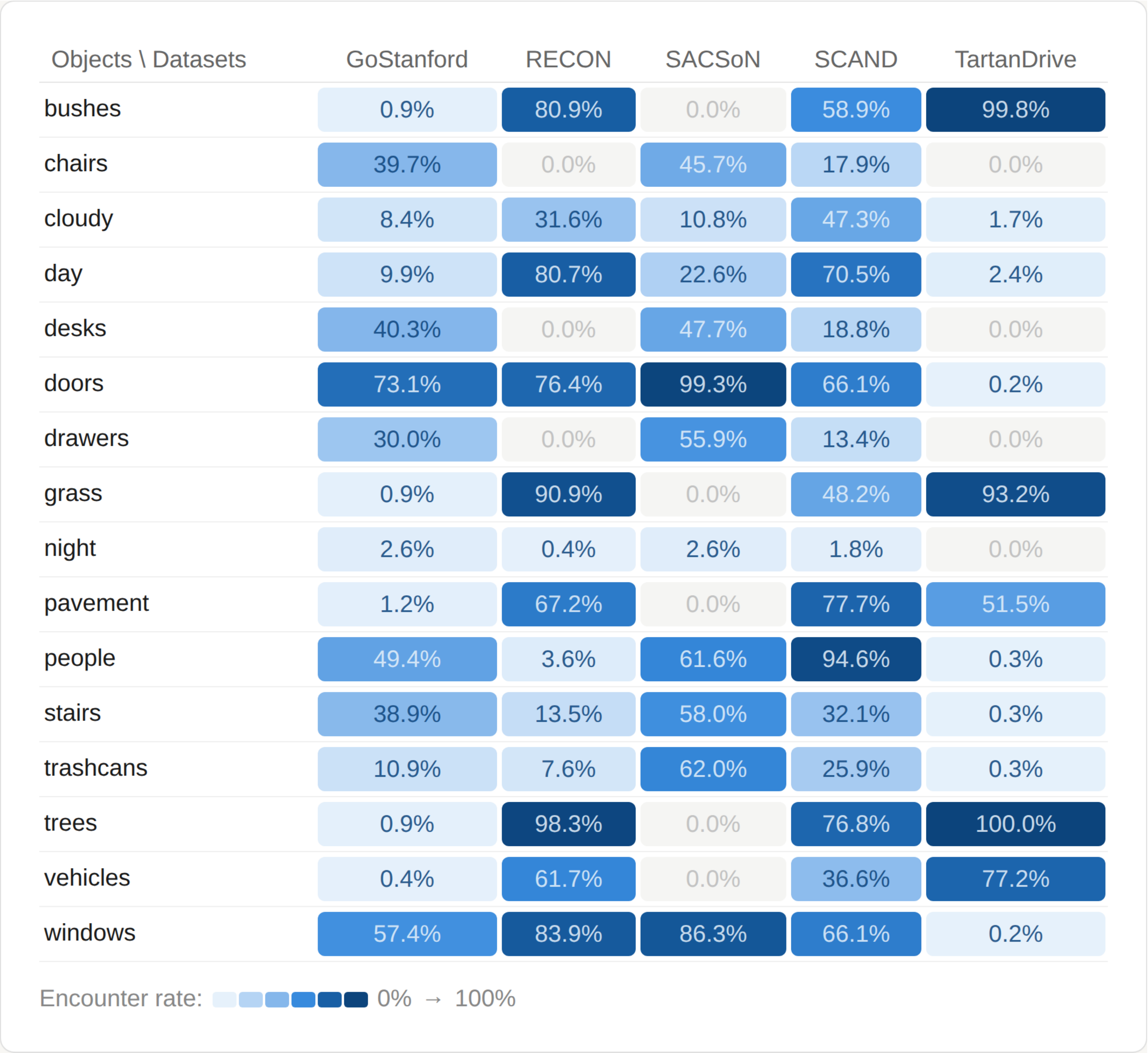}
    \caption{Object encounter probabilities per trajectory across datasets: How likely am I to encounter this object when sampling a trajectory from this dataset?}
    \label{fig:object_encounter}
\end{figure}

We have chosen environment breakdown (See Figure~\ref{fig:dataset_composition} \textit{right}) to reflect the type of environment that are seen during our zero-shot deployment, as well as based on the datasets descriptions when given. Namely, we narrow down the datasets environments as follows: for \textbf{indoor} we have office, hallway, gymnasium, library, and other. As for \textbf{outdoor}, we have forest, park, neighborhood, sidewalk, and others. As for the objects within the environments, we focus on objects that are present in our settings, such as trees, desks, or chairs. See Figure~\ref{fig:object_encounter} for a detailed object composition for each datasets. We also report in Table~\ref{tab:datasets_overview_parameters} datasets' specific information, which includes the total temporal duration hours ($h$), trajectory count, sampling frequency (Hz), scene type (indoor/outdoor), and the average step size.

\textbf{From training Data to Real-World Deployment.} The training datasets indoor versus outdoor ratio contains more outdoor trajectories than indoor (see Figure~\ref{fig:dataset_composition} \textit{left}), more precisely, the outdoor data mostly focuses on neighborhood scenes which consist of grass garden, houses, and trees. However, the outdoor datasets also contain parks and forests, which implies the presence of bushes, trees, and leaves as well. Bushes have high appearance percentage on RECON(80.9\%), SCAND(58.9\%), and Tartan Drive(99.8\%), as do grass and trees (see Figure~\ref{fig:object_encounter}). Those features are present in the Outdoor and forest environments. As for indoor environments, the most prominent settings are the hallway and the office, which are also mainly observed in the Office-Lab environment and the Loop experiments. Overall, doors are seen across all datasets at a high percentage, windows are mostly observed in Go Stanford (57.4\%), RECON(83.9\%), SACSoN(86.3\%) and SCAND(66.1\%). However, chairs are not seen as often 39.7\% for Go Stanford and 45.7\% SACSoN encounter rate. They are widely present in the Office-Lab setting; however, in this environment, they are on the side and not directly in the robot's path.

\begin{table}[t!]
\small
\centering
\caption{\textbf{Characterization of navigation datasets}. Parameters include total temporal duration hours ($h$), trajectory count, sampling frequency (Hz), and scene type (indoor/outdoor). The average step size represents the average Euclidean distance between sequential waypoints at their standard frequencies, 3Hz for GoStanford and 4Hz for the rest.}
\setlength{\tabcolsep}{3.2pt}
\begin{tabular}{@{}lccccc@{}}
\toprule
\textbf{Dataset} & \textbf{Scenes} & \textbf{Trajectories} & \makecell{\textbf{Sampling}\\\textbf{Freq.(Hz)}} & \makecell{\textbf{Average}\\\textbf{Step size}} & \textbf{Duration (h)}\\
\midrule
\rowcolor[gray]{.95} GoStanford  & Indoor   & 3,696  & 3.0\textasciitilde 1.5 & 0.11  & 18 \\
\rowcolor[gray]{.95} SACSoN      & Indoor  & 576     & 12.0\textasciitilde 2.0 &  0.26 & 16 \\
RECON       & Outdoor  & 11,835  & 4.0\textasciitilde 2.0 & 0.24 & 42 \\
SCAND       & Outdoor   & 112     & 12.0\textasciitilde 2.0 & 0.36 & 6 \\
TartanDrive & Outdoor   & 697     & 4.0\textasciitilde 2.0 & 0.81 & 4 \\
\bottomrule
\end{tabular}
\label{tab:datasets_overview_parameters}
\end{table}

\begin{figure}[t!]
    \centering
    \begin{subfigure}[b]{0.32\textwidth}
        \centering
        \includegraphics[width=\textwidth]{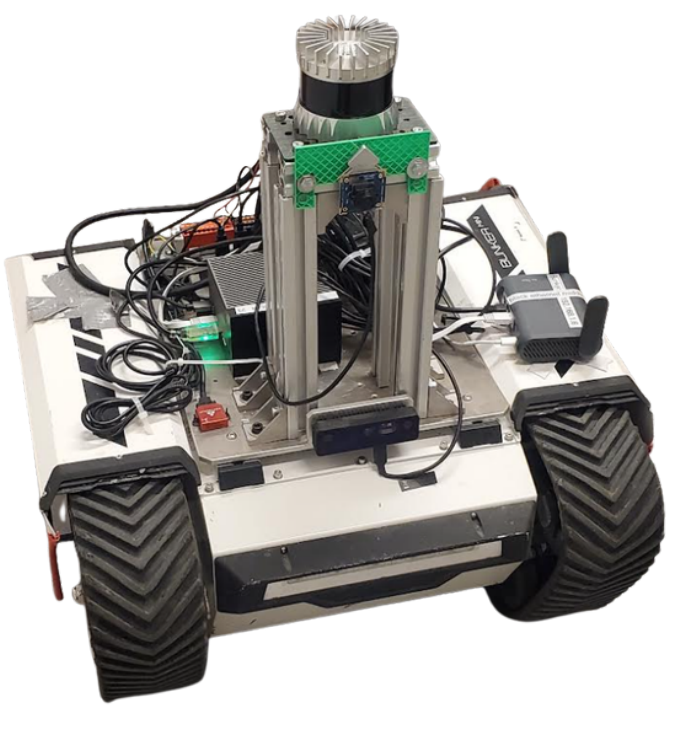} 
        \caption{Bunker - Rover}
        \label{fig:viz_bunker}
    \end{subfigure}
    \hfill 
    \begin{subfigure}[b]{0.20\textwidth}
        \centering
        \includegraphics[width=\textwidth]{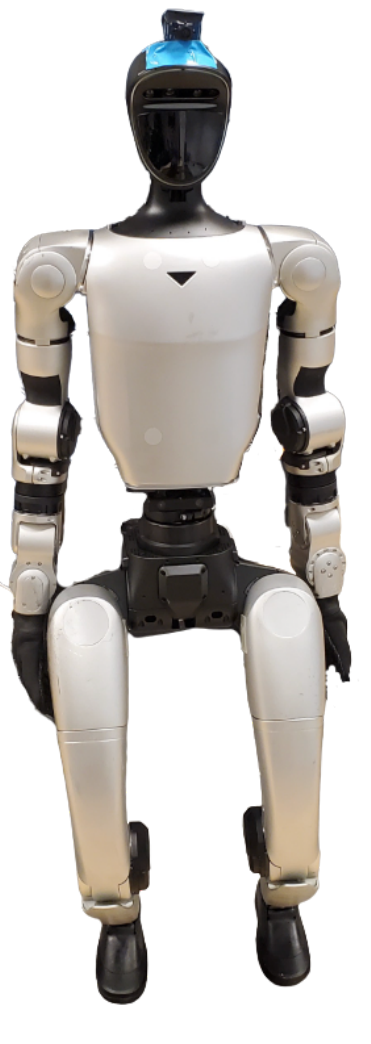} 
        \caption{G1 - Humanoid}
        \label{fig:viz_g1}
    \end{subfigure}
    \hfill 
    \begin{subfigure}[b]{0.32\textwidth}
        \centering
        \includegraphics[width=\textwidth]{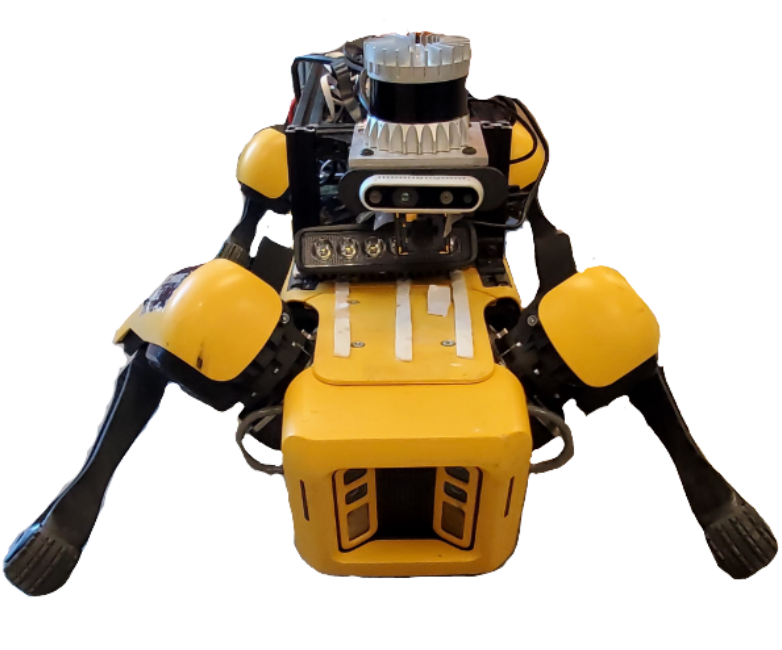} 
        \caption{Spot - Quadruped}
        \label{fig:viz_spot}
    \end{subfigure}
    
    \caption{Visualization of the robots.}
    \label{fig:viz_robots}
\end{figure}

\begin{figure}[b!]
    \centering
    \begin{subfigure}[b]{0.20\textwidth}
        \centering
        \includegraphics[width=\textwidth]{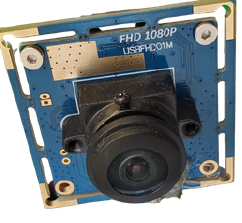} 
        \caption{Fisheye Camera}
        \label{fig:fisheye}
    \end{subfigure}
    \hfill 
    \begin{subfigure}[b]{0.20\textwidth}
        \centering
        \includegraphics[width=\textwidth]{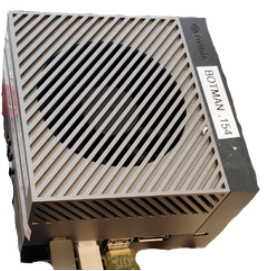} 
        \caption{Jetson Orin Nvidia}
        \label{fig:jetson}
    \end{subfigure}
    \hfill 
    \begin{subfigure}[b]{0.20\textwidth}
        \centering
        \includegraphics[width=\textwidth]{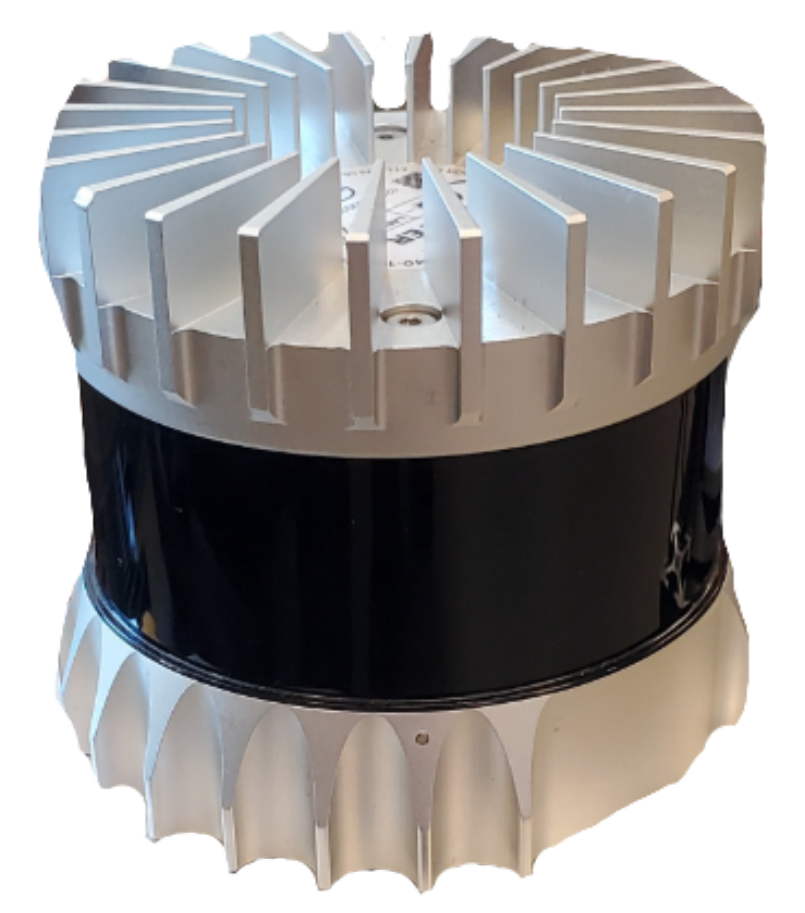} 
        \caption{Ouster 3D LiDAR}
        \label{fig:ousterlidar}
    \end{subfigure}
    
    \caption{Visualization of the onboard sensors.}
    \label{fig:viz_sensors}
\end{figure}

\begin{figure}[t!]
    \centering
    \begin{subfigure}[t]{0.48\textwidth}
        \centering
        \includegraphics[width=\textwidth]{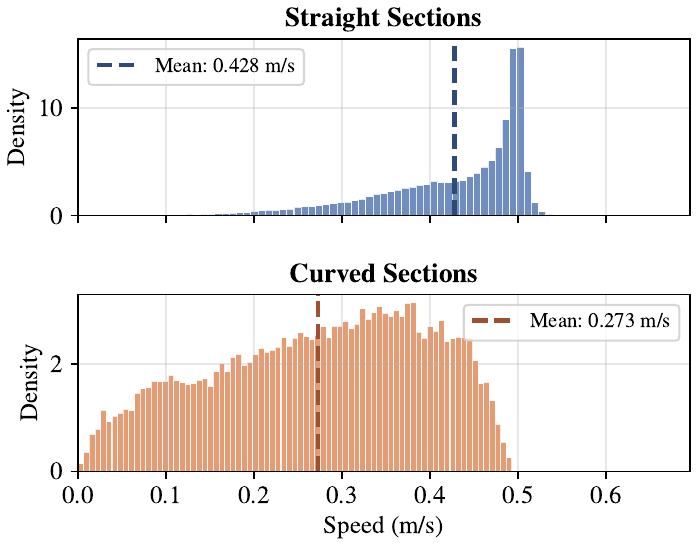}
        \caption{GO Stanford.}
        \label{fig:datasets_speed_go_stanford_speed}
    \end{subfigure}
    \hfill 
    \begin{subfigure}[t]{0.48\textwidth}
        \centering
        \includegraphics[width=\textwidth]{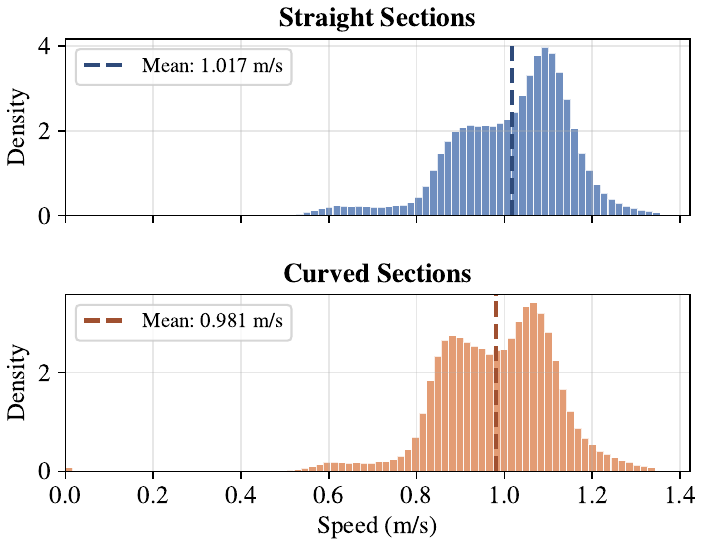}
        \caption{RECON.}
        \label{fig:datasets_speed_recon}
    \end{subfigure}
    
    \begin{subfigure}[t]{0.48\textwidth}
        \centering
        \includegraphics[width=\textwidth]{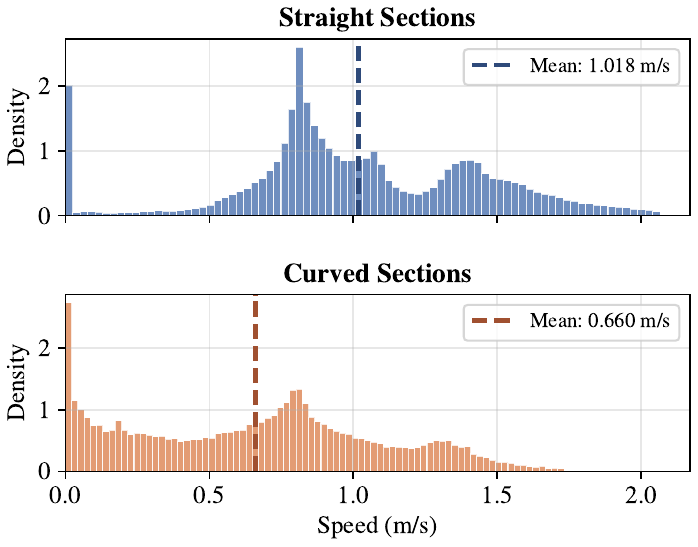}
        \caption{SACSoN.}
        \label{fig:datasets_speed_sacson}
    \end{subfigure}
    \hfill
    \begin{subfigure}[t]{0.48\textwidth}
        \centering
        \includegraphics[width=\textwidth]{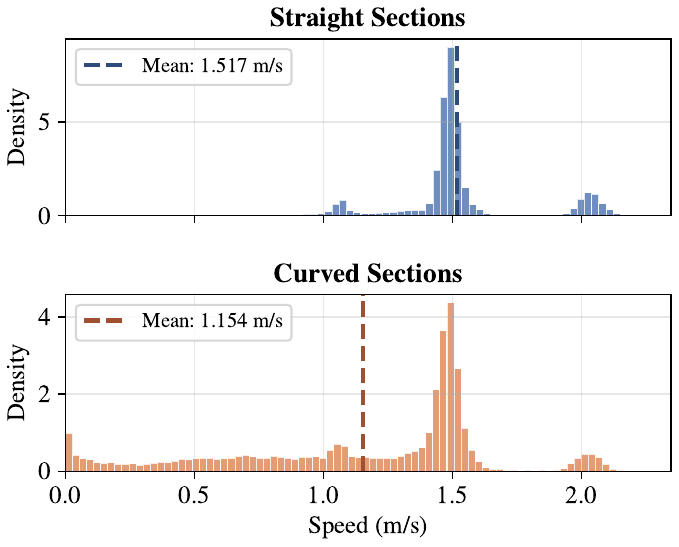}
        \caption{SCAND.}
        \label{fig:datasets_speed_scand}
    \end{subfigure}

    \begin{subfigure}[t]{0.48\textwidth}
        \centering
        \includegraphics[width=\textwidth]{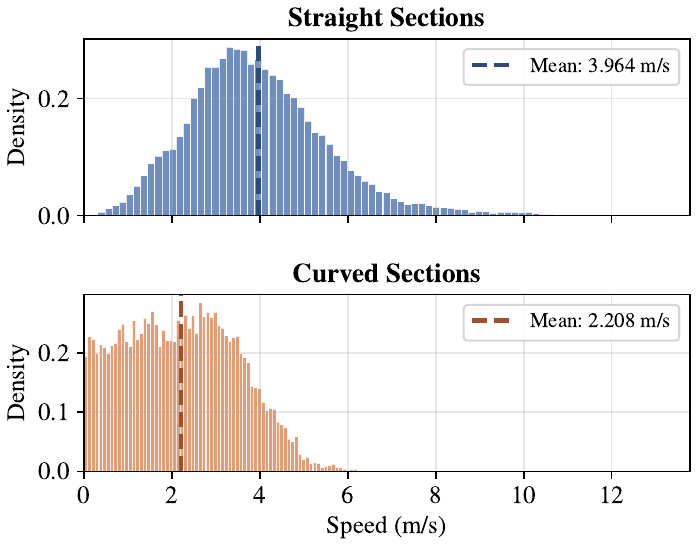}
        \caption{Tartan Drive.}
        \label{fig:datasets_speed_tartan_drive}
    \end{subfigure}
    
    \caption{Speed distributions during straight and curved path sections. Dashed lines indicate the mean speeds.}
    \label{fig:datasets_speed}
\end{figure}

\textbf{Curvature Based Analysis. } Figure~\ref{fig:datasets_speed} presents the results of our curvature-based speed analysis across the datasets. We first smooth each trajectory using a Savitzky--Golay filter with a window size of 11 and a polynomial order of 3 to suppress local noise. We then compute the curvature at each trajectory point and classify high-curvature regions as curved segments and the remaining regions as straight segments, using the 75th percentile of curvature as the threshold.

The resulting speed distributions show that trajectories generally exhibit lower speeds in curved segments than in straight segments, with this trend being particularly pronounced in indoor datasets. These findings suggest that speed variations are strongly correlated with local trajectory geometry, highlighting the importance of distinguishing whether the current motion scale reflects a sustained deployment speed or a temporary slowdown caused by curved or constrained navigation.

\subsubsection{Robot setup}
\label{sec:supp_robot}

Figure~\ref{fig:viz_robots} showcases the robot used for deploying the baselines and our model. Figure~\ref{fig:viz_sensors} shows the main sensors that are relevant to our deployment setup. Tables~\ref{tab:ouster_specs} and~\ref{tab:fisheye_specs} report some of the specifications regarding the LiDAR as well as the fisheye camera, respectively. 

The baselines are deployed onboard on an Agilex Bunker (see Figure~\ref{fig:viz_bunker}) equipped with Nvidia AGX Orin(see Figure~\ref{fig:jetson}). The reference trajectory is recorded by piloting the Bunker manually. The RGB images are recorded using an onboard fisheye camera (see Figure~\ref{fig:fisheye} and Table~\ref{tab:fisheye_specs}), and the poses are given by SuperOdometry(LiDAR)~\citep{zhao2021super} (see Figure~\ref{fig:ousterlidar} and Table~\ref{tab:ouster_specs}).
The time-laspse shown in Figure~\ref{fig:pitch} was deployed on all the robots shown in Figure~\ref{fig:viz_robots}. The G1 humanoid (see Figure~\ref{fig:viz_g1}) does not have an Ouster LiDAR. All of the deployment were done using the Robot Operating System (ROS) software stack - ROS 2. The Spot (see Figure~\ref{fig:viz_spot}) uses an onboard collision avoidance detector; the result shown in the time-lapse of Figure~\ref{fig:pitch} focuses on path following, where collision avoidance is handled by the collision avoidance controller while climbing up the stairs. 

The baselines are exported to ONNX and deployed on board. However, note that the Denoising Diffusion Probabilistic Models Scheduler of NoMaD~\citep{nomad} is not fully converted into ONNX, as some dependencies were required; we converted into ONNX the vision encoder, the distance prediction head, and the noise prediction head.

\begin{table}[t!]
\centering
\caption{LiDAR - Specifications}
\label{tab:ouster_specs}
\begin{tabular}{ll}
\toprule
\textbf{Parameter} & \textbf{Details} \\
\midrule
Product Model & Ouster OS1 \\
Range & Medium Range \\
Resolution & 128 vertical beams \\
Onboard IMU & 3-axis gyro, 3-axis accelerometer \\
\bottomrule
\end{tabular}
\end{table}

\begin{table}[t!]
    \centering
    \caption{Fisheye Camera - Specifications}
    \label{tab:fisheye_specs}
    \small
    \begin{tabular}{ll}
        \toprule
        \textbf{Parameter} & \textbf{Specification} \\
        \midrule
        Model & ELP-USBFHD01M-L180 \\
        Image Sensor & OmniVision OV2710 (1/2.7'' CMOS) \\
        Max. Resolution & Full HD 1920(H) $\times$ 1080(V) \\
        Compression Formats & MJPEG / YUV2 (YUYV) \\
        Lens Type / FOV & 180 Degree Fisheye (187$^\circ$ Diagonal) \\
        Auto Controls & AEC (Exposure), AEB (White Balance) \\
        \bottomrule
    \end{tabular}
\end{table}

\subsubsection{Baselines configurations}
\label{sec:supp_baseline_config}

The baselines configurations and hyperparameters are reported in Table~\ref{tab:parameters_configuration_table}. The model predicts a trajectory of future waypoints for a given subgoal. Rather than making the robot moves toward the nearest predicted waypoint, this index selects which waypoint along that future trajectory to use as an immediate navigation target. A value of 2 means the robot is commanded toward the 3rd predicted waypoint. The distance threshold determines whether the robot has arrived at a topological map node (subgoal node).  When the minimum predicted distance to a node falls below the value of 10, the node is considered reached, and the subgoal node advances to the next node in the topological map. 

At each control step, the subgoal candidate selections does not compare against the entire topological map. Instead it evaluates only nodes in the window $[\,\text{closest\_node} - \text{radius},\ \text{closest\_node} + \text{radius}\,]$.
The number of sample correspond to the number of action trajectories sampled from the model's stochastic action decoder (diffusion policies baselines (NoMaD, MetricNet)). The index of the target node in the topological map image sequence. A value of $-1$ is resolved at load time to the last node in the map, making end-to-end traversal the default behavior. Navigation terminates when $\text{closest\_node} = \text{goal\_node}$, at which point a Bool(True) is published to the ROS2 topic $\text{/topoplan/reached\_goal}$.

\begin{table}[b!]
\centering
\small
\caption{\textbf{Experimental configurations and baseline hyperparameters.} All baselines operated at a fixed frame rate of 4 Hz with maximum linear velocity $v_{\text{max}} = 0.4\text{ m/s}$ and angular velocity $\omega_{\text{max}} = 1.2\text{ rad/s}$. Parameters $w, t, r,$ and $n$ define the waypoint selection and sampling logic.}
\label{tab:parameters}
\begin{tabular}{@{}l c c@{}}
\toprule
\textbf{Parameter} & \textbf{ViNT, NoMaD, MetricNet} & \textbf{VISTA, VISTA w/o AH} \\ \midrule
Waypoint Index ($w$)        & 2             & 1              \\
Close Threshold ($t$)       & 10            & 10             \\
Look-ahead Radius ($r$)     & 4             & 2              \\
Number of Samples ($n$)     & 8             & 8              \\
Goal Node Index ($g$)       & $-1^{\dagger}$           & $-1^{\dagger}$ \\ 
\midrule
\rowcolor[gray]{.95} \multicolumn{3}{l}{\textit{Shared Global Constraints}} \\
Max. Linear Vel. ($v_{\text{max}}$)  & \multicolumn{2}{c}{0.4 m/s} \\
Max. Angular Vel. ($\omega_{\text{max}}$) & \multicolumn{2}{c}{1.2 rad/s} \\
Frame Rate                  & \multicolumn{2}{c}{4 Hz}    \\
\bottomrule
\addlinespace[1ex]
\multicolumn{3}{l}{\footnotesize $^{\dagger}$Indicates the terminal node in the topological map.}
\end{tabular}
\label{tab:parameters_configuration_table}
\end{table}

\subsection{Additional Results and Analyses}
\label{sec:supp_experiments}

\subsubsection{Supplementary Note: Offline scale-consistency evaluation}
\label{sec:supp_scale_consistency}

Since scale consistency only becomes a problem when there is a curved path, we applied a curvature-based trajectory segmentation, then used only the subset of the dataset where the target trajectory belonged to a curved section. The subgoal image was selected as the image 20 steps ahead of the current observation image, to fully cover the prediction horizon of 5 in the spacing 4 case ($5\times 4 = 20$). The prediction of NoMaD was truncated to 5 steps to match the prediction horizon of other models.

\subsubsection{Environments in detail}
\label{sec:supp_environments}

\begin{figure}[t]
    \centering
  \includegraphics[width=1.\textwidth]{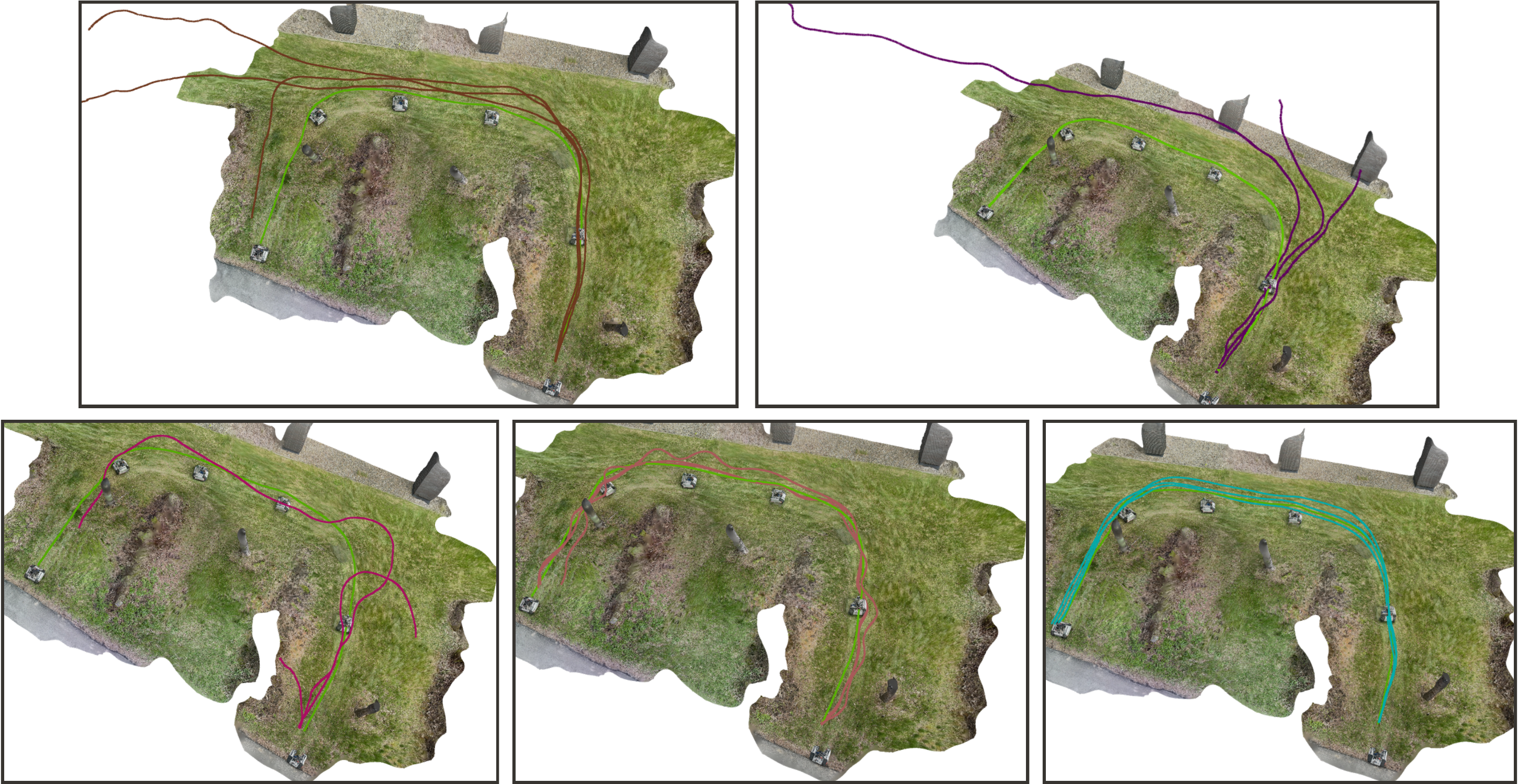}
  \caption{\prettyvint{}, \prettynomad{}, \prettymetnet{}, \prettydino{} and the \prettyablation{} for all (3 trials) outdoor deployment results with the \prettyref{} trajectory. Visualization is qualitative as trajectory-to-map alignment is approximate. See Table~\ref{tab:outcomes_sr_coll_inc_outdoor_forest_lab} and Figure~\ref{fig:real_world_pathlen_navpercent} for qualitative results.}
  \label{fig:outdoor_baselines}
\end{figure}

\textbf{Grass Field Structured Environment (Outdoor)}
Evaluations were conducted on a grass field (see Figure~\ref{fig:outdoor_baselines}) featuring uneven terrain, gravel, and obstacles (concrete pillars, trees, and dense bushes). The environment presents challenges characteristic of unstructured outdoor navigation: absence of distinct features, spatially repetitive visual texture, and moderate terrain elevation. 

\begin{figure}[t]
    \centering
    \includegraphics[width=\linewidth]{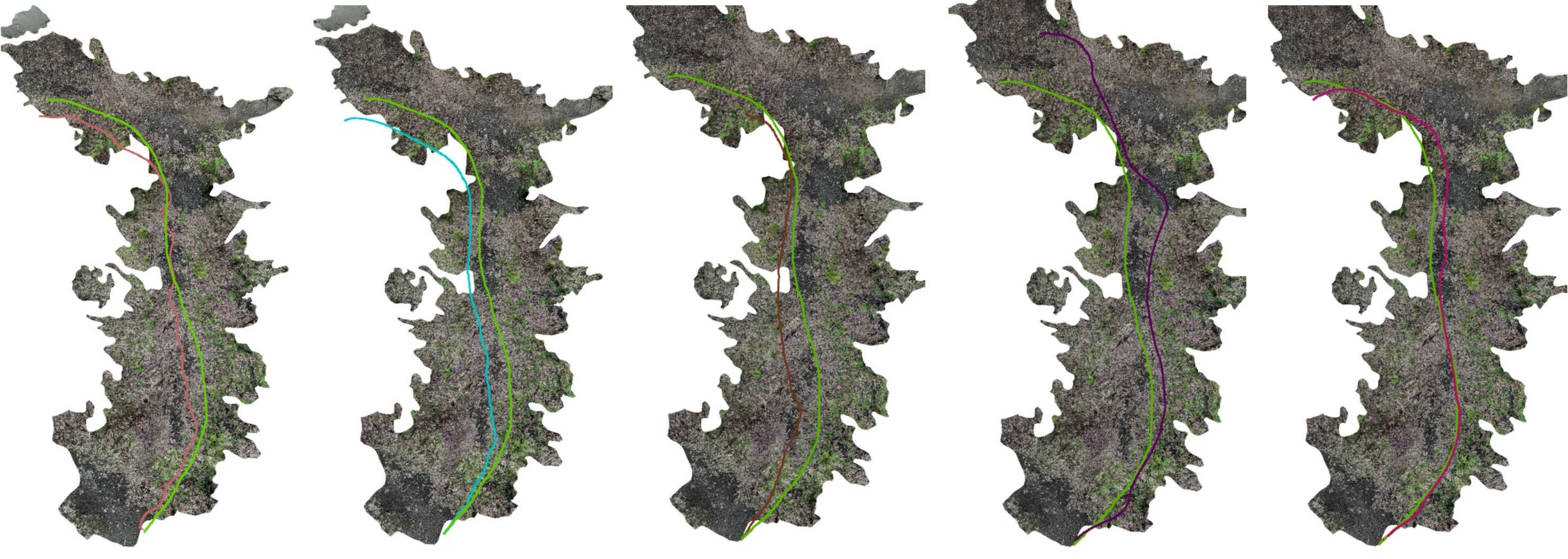}
    \caption{Forest setting: For clarity, we visualize a single illustrative trial per method; quantitative results across all trials are reported in Table~\ref{tab:outcomes_sr_coll_inc_outdoor_forest_lab} and Figure~\ref{fig:real_world_pathlen_navpercent}. Visualization is qualitative as trajectory-to-map alignment is approximate. We show \prettydino{}, \prettyablation{}, \prettyvint{}, \prettynomad{} and \prettymetnet{} with \prettyref{} path.}
    \label{fig:forest_mesh_fig}
\end{figure}

\textbf{Unstructured Natural Environment (Forest)}
A forested trail (see Figure~\ref{fig:forest_mesh_fig}) environment with dense, visually similar vegetation, irregular ground cover of leaves, and a lack of explicit path delineation. In this setting, the ambiguity comes from the near-uniform appearance of the forest floor, which provides minimal discriminative features, whereas richer features are given by the trees and bushes. Navigation in this environment requires the model to focus on the vertical features in the images rather than the ground plane. 

\begin{figure}[t]
    \centering
    \includegraphics[width=\linewidth]{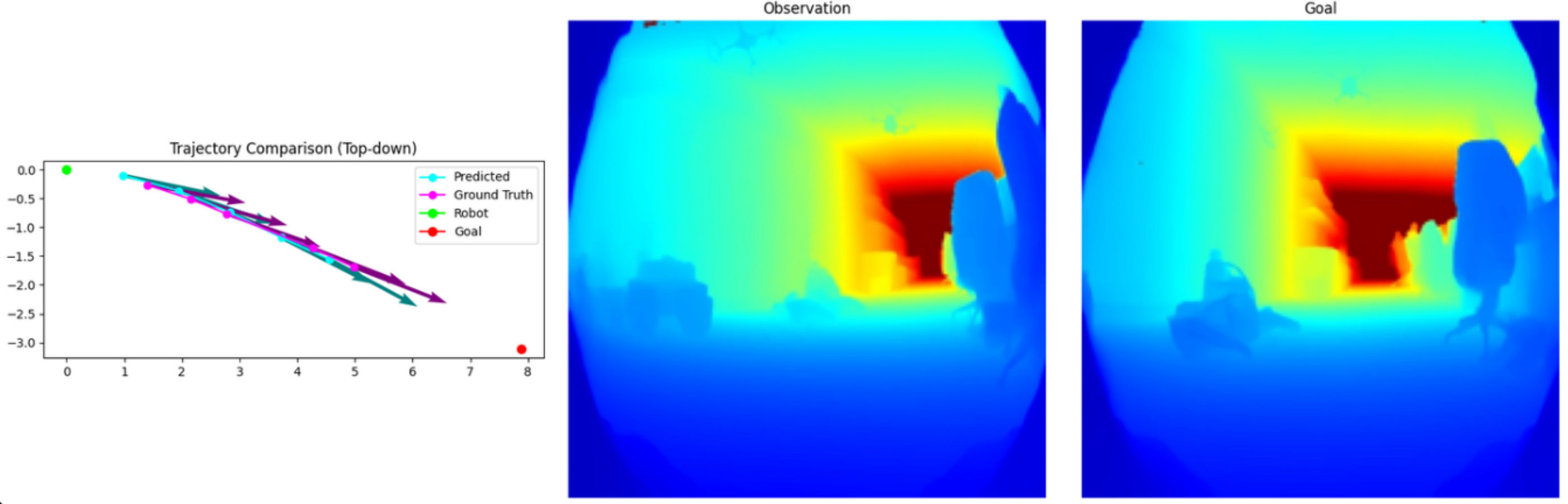}
    \includegraphics[width=\linewidth]{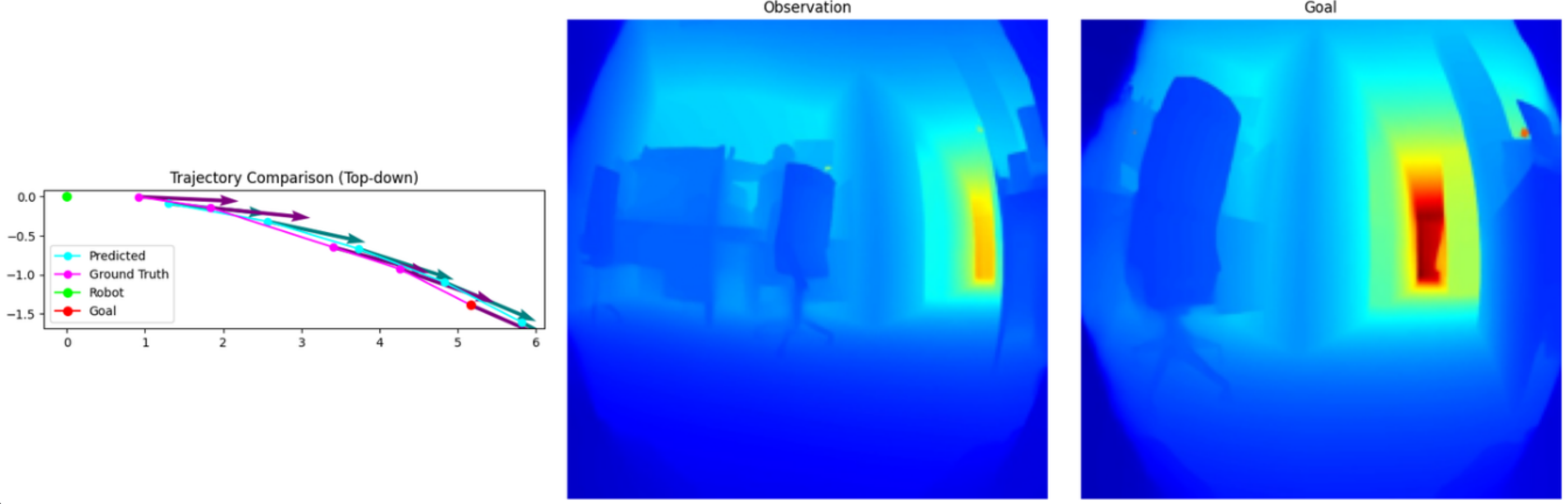}
    \caption{VISTA depth predictions visualization}
    \label{fig:VISTA_depth_sampled}
\end{figure}

\begin{figure}[t]
    \centering
    \begin{subfigure}{\linewidth}
        \centering
        \includegraphics[width=\linewidth]{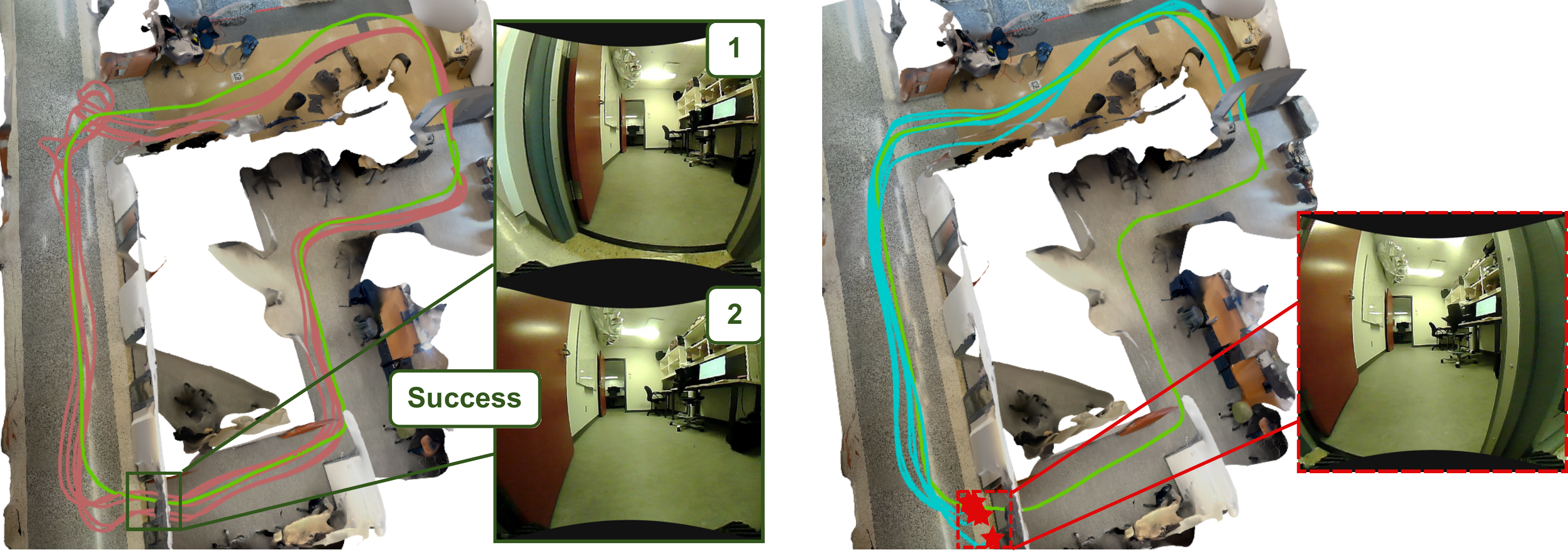}
        \caption{\prettydino{} and \prettyablation{}.}
        \label{fig:lab_ours_fig}
    \end{subfigure}
    
    \begin{subfigure}{\linewidth}
        \centering
        \includegraphics[width=\linewidth]{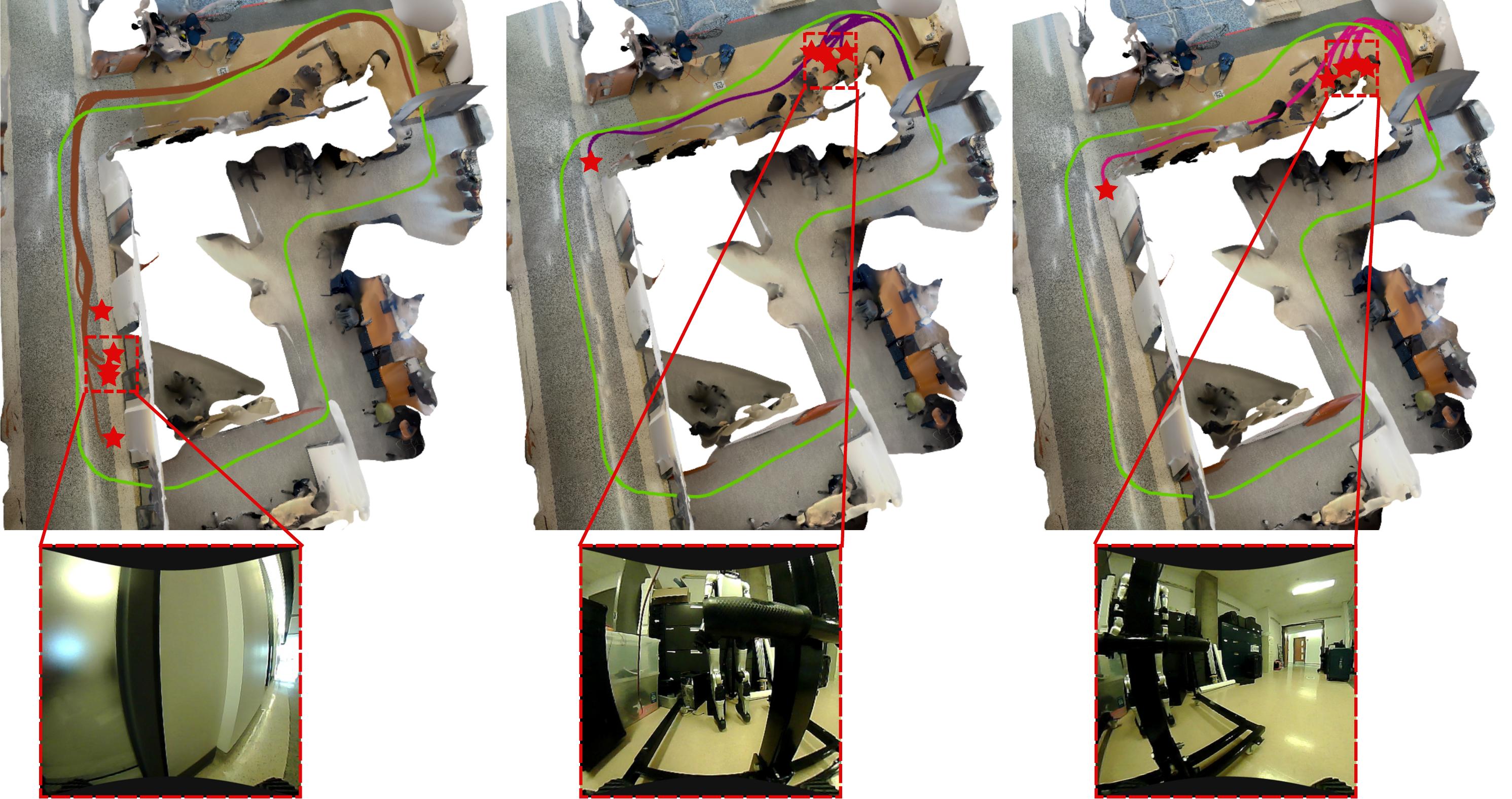}
        \caption{\prettyvint{}, \prettynomad{} and \prettymetnet{}.}
        \label{fig:lab_baselines_fig}
    \end{subfigure}

    \caption{We report all trials (5) for all baselines with the \prettyref{} trajectory. We highlight collisions with red stars. Visualization is qualitative as trajectory-to-map alignment is approximate, quantitative results can be found in Table~\ref{tab:outcomes_sr_coll_inc_outdoor_forest_lab} and Figure~\ref{fig:real_world_pathlen_navpercent}.}
    \label{fig:lab_meshes_fig}
\end{figure}

\textbf{Indoor Cluttered Environment (Office-Lab)}
The indoor (see Figure~\ref{fig:lab_meshes_fig}) evaluation environment consists of a multi-room laboratory space connected by narrow doorways, with clutter including equipment racks, furniture, and movable obstacles. The environment demands precise control to navigate tight doorframes as well as close path-following to avoid obstacles. This setting specifically stress-tests the model's ability to generalize spatial reasoning in cluttered scenario.

\begin{figure}[b]
    \centering
    \includegraphics[width=\linewidth]{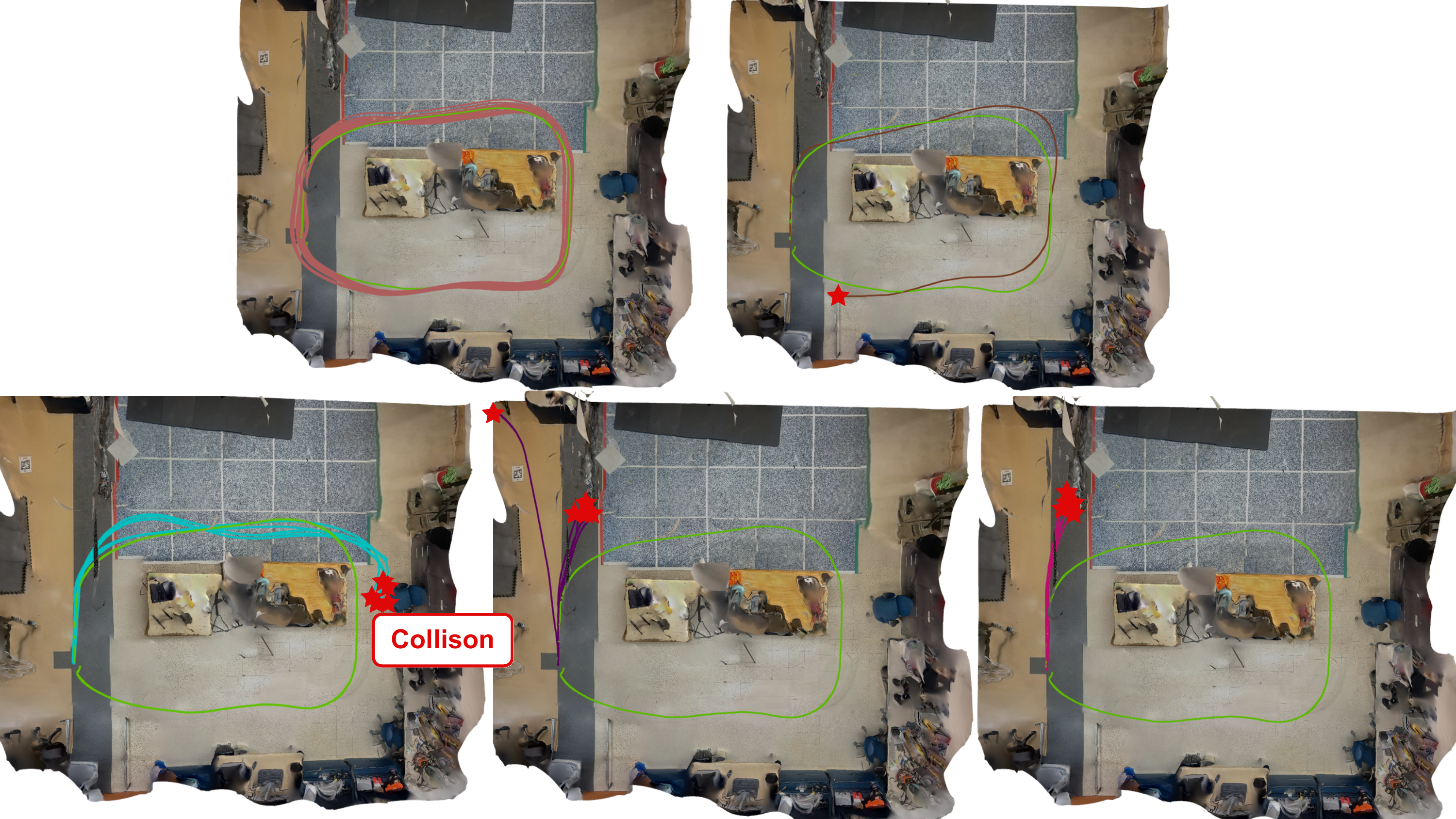}
    \caption{Visual representation of the loop experiments with \prettydino{}, \prettyvint{}, \prettyablation{}, \prettynomad{} and \prettymetnet{} with \prettyref{} path. For visibility, we report one trial for \prettyvint{}. We highlight collisions with red stars.  Note that the figure is qualitative as trajectory-to-map alignment is approximate. For quantitative results, see Figure~\ref{fig:path_length_by_env_loop5}.}
    \label{fig:loop5_fig}
\end{figure}

\begin{figure}[b]
    \centering
    \includegraphics[width=0.65\textwidth]{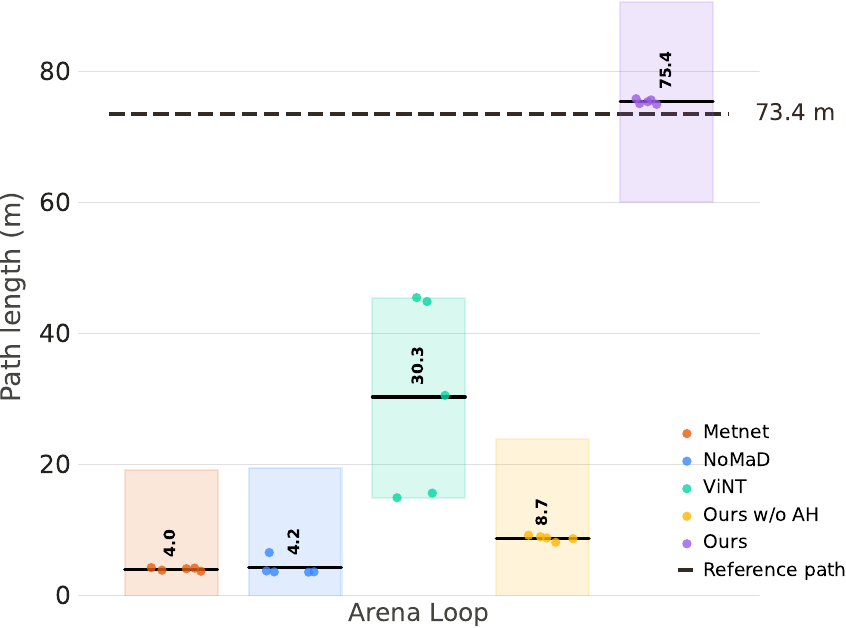}
    \caption{\textbf{Loop robustness for topological navigation.} For a given topological map corresponding to a loop, each baseline is tasked to traverse the loop 5 times. Accumulated distance prediction errors hinder sub-goal selection, leading to collision or getting stuck.}
    \label{fig:path_length_by_env_loop5}
\end{figure}

\subsubsection{Robustness to deployment step size}
\label{sec:supp_stepsize_robustness}

\begin{figure}[b]
    \centering
    \begin{subfigure}{\linewidth}
        \centering
        \includegraphics[width=\linewidth]{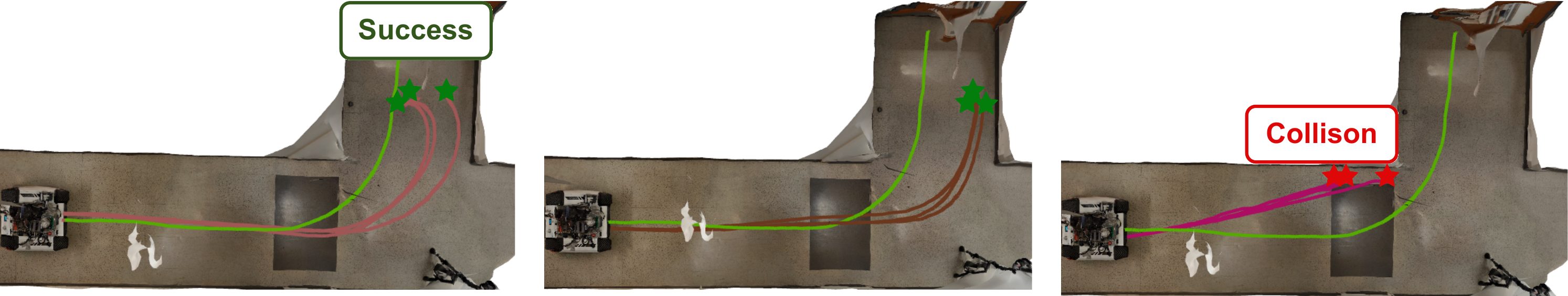}
        \caption{Default trajectory configuration.}
        \label{fig:sharpturn_default}
    \end{subfigure}
    
    \begin{subfigure}{\linewidth}
        \centering
        \includegraphics[width=\linewidth]{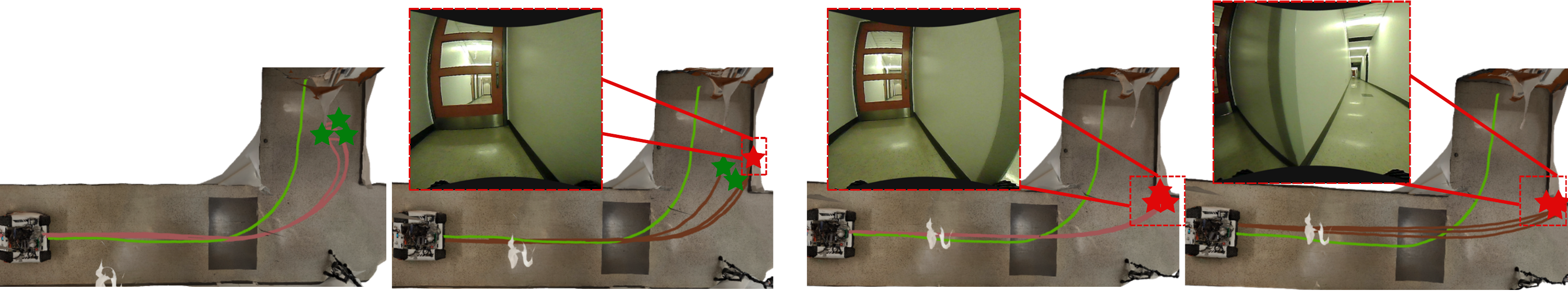}
        \caption{2 Hz (\textit{left} ) and 1 Hz (\textit{right}).}
        \label{fig:sharpturn_hz}
    \end{subfigure}
        
    \begin{subfigure}{\linewidth}
        \centering
        \includegraphics[width=\linewidth]{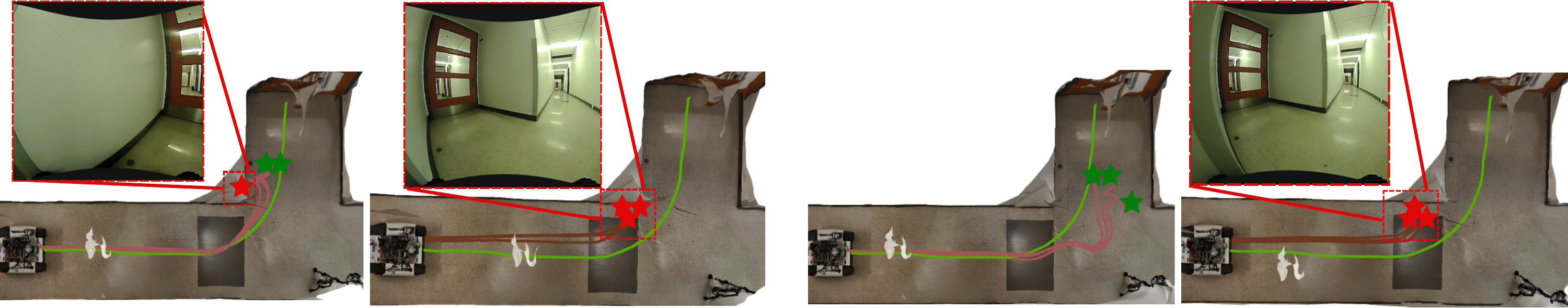}
        \caption{0.2 m/s (\textit{left} ) and 0.1 m/s (\textit{right}).}
        \label{fig:sharpturn_ms}
    \end{subfigure}
    
    \caption{Changes of control frequency and maximum linear speed. Visualization is qualitative as trajectory-to-map alignment is approximate. The reported baselines are: \prettydino{}, \prettyvint{} and \prettymetnet{} with the \prettyref{} trajectory. We highlight collisions with red stars and successes with green.}
    \label{fig:all_sharpturn_fig}
\end{figure}

We showcase the change in control frequency and maximum linear speed in Figure~\ref{fig:all_sharpturn_fig} as a qualitative visualization. Figure~\ref{fig:sharpturn_default} reports the default configuration, which corresponds to a control frequency of 4 Hz and a maximum linear speed of 0.4 m/s for the following baseline: \prettydino{}, \prettyvint{} and \prettymetnet{}. Figure~\ref{fig:sharpturn_hz} reports the experiments with the changes in control frequency, whereas Figure~\ref{fig:sharpturn_ms} shows the variation in maximum linear speed. The quantitative results are reported in Table~\ref{tab:varied_stepsize_results}.

\subsubsection{Depth prediction results}
\label{sec:supp_depth}

To stress-test the out-of-distribution generalization capacity of VISTA, we evaluated the model on depth images (see Figure~\ref{fig:VISTA_depth_sampled} and Table~\ref{tab:depth_results}), which are entirely absent from the RGB-only training datasets. No fine-tuning or domain adaptation was performed. The evaluation subset datasets were taken from ROS bags collected in the office-lab environment, which features an indoor structure including chairs, desks, computers, doors, and windows.
Despite never having observed depth images during training, VISTA produces waypoint predictions that remain close to the RGB-supervised ground truth, suggesting that the DINOv3 backbone extracts geometric and structural representations sufficiently general to transfer across different image modalities. We hypothesize that those results indicate that the model's learned features can capture scene geometry, enabling zero-shot depth generalization.

\subsubsection{Additional Experiments; The loop}
\label{sec:supp_loop}

In these experiments, we are testing out the reliability of the baseline's distance predictions, as they are used for subgoal selection. Accumulated error would cause a wrong subgoal, potentially leading to a collision. To test this, we collected a single loop topological map within a workshop-like setting (see Figure~\ref{fig:loop5_fig}). The baselines need to perform 5 loops for a total of 5 trials. We showcase in Figure~\ref{fig:loop5_fig} a qualitative visual representation of the collisions points within that environment. We can observe that VISTA consistently achieves the loop reliably as shown in Figure~\ref{fig:loop5_fig} and in Figure~\ref{fig:path_length_by_env_loop5}, where the path length of VISTA remains close to the path length of the original loop multiplied by 5 (for the five loop).

\begin{algorithm}[b]
\DontPrintSemicolon
\caption{Closest Node, Waypoint Selection for Topological Navigation}
\label{alg:deploy_algo}

\KwIn{%
  topological map $M$,\;
  goal node index $g_{node}$ \texttt{(args.goal, user defined)},\;
  search radius $r$ \texttt{(args.radius, user defined)},\;
  proximity threshold $\tau$ \texttt{(args.close\_threshold, user defined)},\;
  waypoint index $w_{idx}$ \texttt{(args.waypoint, user defined)}%
}
\KwOut{%
  \texttt{chosen\_waypoint} \mycomment{navigation target waypoint at each time step}%
}

\BlankLine
\tcp{Initialisation}
$n_c \leftarrow 0$ \mycomment{closest node starts at first topological map node}\;

\BlankLine
\While{\upshape robot has not reached goal}{

  \BlankLine
  \tcp{Step 1 — Compute local moving window}
  $start \leftarrow \max(n_c - r,\ 0)$\;
  $end \leftarrow \min(n_c + r + 1,\ g_{node})$\;

  \BlankLine
  \tcp{Step 2 — Build image batches}
  $B_\text{obs} \leftarrow [\,]$,\quad $B_\text{goal} \leftarrow [\,]$\;
  \For{$i \leftarrow start$ \KwTo $end$}{
    $B_\text{obs}  \mathrel{+}= \textsc{TransformImage}(\text{observation history})$\;
    $B_\text{goal} \mathrel{+}= \textsc{TransformImage}(\mathcal{M}[i])$\;
  }

  \BlankLine
  \tcp{Step 3 — Run inference model}
  $\mathbf{dists},\ \mathbf{wps} \leftarrow \textsc{Model}(B_\text{obs},\ B_\text{goal})$
  \mycomment{distances and waypoints per $B_\text{goal}$ batches}\;

  \BlankLine
  \tcp{Step 4 — Closest node belief}
  $d_{min}^{idx} \leftarrow \arg\min_{j}\ \mathbf{dists}[j]$\;

  \BlankLine
  \tcp{Step 5 — Update closest node and select waypoint}
  \eIf{$\mathbf{dists}[d_{min}^{idx}] > \tau$}{
    \mycomment{Robot has not yet reached node $d_{min}^{idx}$ - stay at current node}\;
    $\texttt{chosen\_waypoint} \leftarrow \mathbf{wps}[d_{min}^{idx}][w_{idx}]$\;
    $n_c \leftarrow start + d_{min}^{idx}$\;
  }{
    \mycomment{Robot is already close to subgoal node - advance to next node subgoal}\;
    $\texttt{chosen\_waypoint} \leftarrow \mathbf{wps}\!\left[\min(d_{min}^{idx}+1,\,|\mathbf{wps}|-1)\right][w_{idx}]$\;
    $n_c \leftarrow \min(start + d_{min}^{idx} + 1,\ g_{node})$\;
  }

  \BlankLine
  \tcp{Step 6 — ROS\footnote{Robot Operating System} Publish and navigate}
  \textsc{Publish}(\texttt{chosen\_waypoint})\;

    \eIf{$n_c = g_{node}$}{
      Goal node is reached\;
    }{
    continue while loop
    }
  
}

\end{algorithm}


\begin{thebibliography}{27}
\providecommand{\natexlab}[1]{#1}
\providecommand{\url}[1]{\texttt{#1}}
\expandafter\ifx\csname urlstyle\endcsname\relax
  \providecommand{\doi}[1]{doi: #1}\else
  \providecommand{\doi}{doi: \begingroup \urlstyle{rm}\Url}\fi

\bibitem[Walke et~al.(2023)Walke, Black, Lee, Kim, Du, Zheng, Zhao, Hansen-Estruch, Vuong, He, Myers, Fang, Finn, and Levine]{walke2024bridgedatav2datasetrobot}
H.~Walke, K.~Black, A.~Lee, M.~J. Kim, M.~Du, C.~Zheng, T.~Zhao, P.~Hansen-Estruch, Q.~Vuong, A.~He, V.~Myers, K.~Fang, C.~Finn, and S.~Levine.
\newblock {BridgeData V2}: {A} {D}ataset for {R}obot {L}earning at {S}cale.
\newblock In \emph{Conf. on Robot Learning ({CoRL})}, 2023.

\bibitem[Khazatsky et~al.(2025)Khazatsky, Pertsch, Nair, Balakrishna, Dasari, Karamcheti, Nasiriany, Srirama, Chen, Ellis, Fagan, Hejna, Itkina, Lepert, Ma, Miller, Wu, Belkhale, Dass, Ha, Jain, Lee, Lee, Memmel, Park, Radosavovic, Wang, Zhan, Black, Chi, Hatch, Lin, Lu, Mercat, Rehman, Sanketi, Sharma, Simpson, Vuong, Walke, Wulfe, Xiao, Yang, Yavary, Zhao, Agia, Baijal, Castro, Chen, Chen, Chung, Drake, Foster, Gao, Guizilini, Herrera, Heo, Hsu, Hu, Irshad, Jackson, Le, Li, Lin, Lin, Ma, Maddukuri, Mirchandani, Morton, Nguyen, O'Neill, Scalise, Seale, Son, Tian, Tran, Wang, Wu, Xie, Yang, Yin, Zhang, Bastani, Berseth, Bohg, Goldberg, Gupta, Gupta, Jayaraman, Lim, Malik, Martín-Martín, Ramamoorthy, Sadigh, Song, Wu, Yip, Zhu, Kollar, Levine, and Finn]{khazatsky2025droidlargescaleinthewildrobot}
A.~Khazatsky, K.~Pertsch, S.~Nair, A.~Balakrishna, S.~Dasari, S.~Karamcheti, S.~Nasiriany, M.~K. Srirama, L.~Y. Chen, K.~Ellis, P.~D. Fagan, J.~Hejna, M.~Itkina, M.~Lepert, Y.~J. Ma, P.~T. Miller, J.~Wu, S.~Belkhale, S.~Dass, H.~Ha, A.~Jain, A.~Lee, Y.~Lee, M.~Memmel, S.~Park, I.~Radosavovic, K.~Wang, A.~Zhan, K.~Black, C.~Chi, K.~B. Hatch, S.~Lin, J.~Lu, J.~Mercat, A.~Rehman, P.~R. Sanketi, A.~Sharma, C.~Simpson, Q.~Vuong, H.~R. Walke, B.~Wulfe, T.~Xiao, J.~H. Yang, A.~Yavary, T.~Z. Zhao, C.~Agia, R.~Baijal, M.~G. Castro, D.~Chen, Q.~Chen, T.~Chung, J.~Drake, E.~P. Foster, J.~Gao, V.~Guizilini, D.~A. Herrera, M.~Heo, K.~Hsu, J.~Hu, M.~Z. Irshad, D.~Jackson, C.~Le, Y.~Li, K.~Lin, R.~Lin, Z.~Ma, A.~Maddukuri, S.~Mirchandani, D.~Morton, T.~Nguyen, A.~O'Neill, R.~Scalise, D.~Seale, V.~Son, S.~Tian, E.~Tran, A.~E. Wang, Y.~Wu, A.~Xie, J.~Yang, P.~Yin, Y.~Zhang, O.~Bastani, G.~Berseth, J.~Bohg, K.~Goldberg, A.~Gupta, A.~Gupta, D.~Jayaraman, J.~J. Lim, J.~Malik, R.~Martín-Martín, S.~Ramamoorthy, D.~Sadigh,
  S.~Song, J.~Wu, M.~C. Yip, Y.~Zhu, T.~Kollar, S.~Levine, and C.~Finn.
\newblock Droid: A large-scale in-the-wild robot manipulation dataset, 2025.
\newblock URL \url{https://arxiv.org/abs/2403.12945}.

\bibitem[Shah et~al.(2023)Shah, Sridhar, Dashora, Stachowicz, Black, Hirose, and Levine]{vint}
D.~Shah, A.~Sridhar, N.~Dashora, K.~Stachowicz, K.~Black, N.~Hirose, and S.~Levine.
\newblock {ViNT}: {A} {F}oundation {M}odel for {V}isual {N}avigation.
\newblock In \emph{Conf. on Robot Learning ({CoRL})}, 2023.

\bibitem[Sridhar et~al.(2024)Sridhar, Shah, Glossop, and Levine]{nomad}
A.~Sridhar, D.~Shah, C.~Glossop, and S.~Levine.
\newblock {NoMaD}: {G}oal {M}asked {D}iffusion {P}olicies for {N}avigation and {E}xploration.
\newblock In \emph{Int. Conf. on Robotics and Automation ({ICRA})}, 2024.

\bibitem[Nayak et~al.(2025)Nayak, Oliveira, Gode, Schmid, and Burgard]{nayak2025metricnet}
A.~Nayak, D.~N. Oliveira, S.~Gode, C.~Schmid, and W.~Burgard.
\newblock Metricnet: Recovering metric scale in generative navigation policies.
\newblock \emph{arXiv preprint arXiv:2509.13965}, 2025.

\bibitem[Guerrier et~al.(2026)Guerrier, Soma, Pavlasek, and Beltrame]{guerrier2026visionfoundationmodelsnavigate}
M.~Guerrier, K.~Soma, J.~Pavlasek, and G.~Beltrame.
\newblock Can vision foundation models navigate? zero-shot real-world evaluation and lessons learned, 2026.
\newblock URL \url{https://arxiv.org/abs/2603.25937}.

\bibitem[Shah et~al.(2023)Shah, Sridhar, Bhorkar, Hirose, and Levine]{gnm}
D.~Shah, A.~Sridhar, A.~Bhorkar, N.~Hirose, and S.~Levine.
\newblock {GNM}: {A} {G}eneral {N}avigation {M}odel to {D}rive {A}ny {R}obot.
\newblock In \emph{Int. Conf. on Robotics and Automation ({ICRA})}, 2023.

\bibitem[Qiao et~al.(2025)Qiao, Lyu, Wang, Wang, Li, Zhang, Tan, and Wu]{opennavQiao}
Y.~Qiao, W.~Lyu, H.~Wang, Z.~Wang, Z.~Li, Y.~Zhang, M.~Tan, and Q.~Wu.
\newblock Open-nav: Exploring zero-shot vision-and-language navigation in continuous environment with open-source llms.
\newblock In \emph{2025 IEEE International Conference on Robotics and Automation (ICRA)}, pages 6710--6717, 2025.
\newblock \doi{10.1109/ICRA55743.2025.11127584}.

\bibitem[Wen et~al.(2025)Wen, Huang, Huang, Huang, Yuan, Hao, Lin, Liu, and Fang]{zeroshot_objNav}
C.~Wen, Y.~Huang, H.~Huang, Y.~Huang, S.~Yuan, Y.~Hao, H.~Lin, Y.-S. Liu, and Y.~Fang.
\newblock Zero-shot object navigation with vision-language models reasoning.
\newblock In A.~Antonacopoulos, S.~Chaudhuri, R.~Chellappa, C.-L. Liu, S.~Bhattacharya, and U.~Pal, editors, \emph{Pattern Recognition}, pages 389--404, Cham, 2025. Springer Nature Switzerland.
\newblock ISBN 978-3-031-78456-9.

\bibitem[Yin et~al.(2025)Yin, Xu, Zhao, Wang, Zhou, and Lu]{Yin_2025_CVPR}
H.~Yin, X.~Xu, L.~Zhao, Z.~Wang, J.~Zhou, and J.~Lu.
\newblock Unigoal: Towards universal zero-shot goal-oriented navigation.
\newblock In \emph{Proceedings of the IEEE/CVF Conference on Computer Vision and Pattern Recognition (CVPR)}, pages 19057--19066, June 2025.

\bibitem[Wang et~al.(2025)Wang, Hu, Tang, and Gao]{coal}
Z.~Wang, J.~Hu, Q.~Tang, and W.~Gao.
\newblock {COAL}: {R}obust {C}ontrastive {L}earning-{B}ased {V}isual {N}avigation {F}ramework.
\newblock \emph{Journal of Field Robotics}, 42\penalty0 (5):\penalty0 2028--2041, 2025.
\newblock \doi{https://doi.org/10.1002/rob.22508}.
\newblock URL \url{https://onlinelibrary.wiley.com/doi/abs/10.1002/rob.22508}.

\bibitem[Bar et~al.(2025)Bar, Zhou, Tran, Darrell, and LeCun]{navworldmodel}
A.~Bar, G.~Zhou, D.~Tran, T.~Darrell, and Y.~LeCun.
\newblock {N}avigation {W}orld {M}odels.
\newblock In \emph{Conf. on Comp. Vision and Pattern Rec. ({CVPR})}, 2025.

\bibitem[Wang et~al.(2024)Wang, Tan, and Nejat]{NavFormer}
H.~Wang, A.~H. Tan, and G.~Nejat.
\newblock {NavFormer}: A transformer architecture for robot target-driven navigation in unknown and dynamic environments.
\newblock \emph{Robotics and Automation Letters}, 2024.
\newblock \doi{10.1109/LRA.2024.3412638}.

\bibitem[{Octo Model Team} et~al.(2024){Octo Model Team}, Ghosh, Walke, Pertsch, Black, Mees, Dasari, Hejna, Xu, Luo, Kreiman, Tan, Chen, Sanketi, Vuong, Xiao, Sadigh, Finn, and Levine]{octo_2023}
{Octo Model Team}, D.~Ghosh, H.~Walke, K.~Pertsch, K.~Black, O.~Mees, S.~Dasari, J.~Hejna, C.~Xu, J.~Luo, T.~Kreiman, Y.~Tan, L.~Y. Chen, P.~Sanketi, Q.~Vuong, T.~Xiao, D.~Sadigh, C.~Finn, and S.~Levine.
\newblock {O}cto: {A}n {O}pen-{S}ource {G}eneralist {R}obot {P}olicy.
\newblock In \emph{Robotics: Science and Systems}, Delft, Netherlands, 2024.

\bibitem[Bharadhwaj et~al.(2024)Bharadhwaj, Vakil, Sharma, Gupta, Tulsiani, and Kumar]{bharadhwaj2023roboagentgeneralizationefficiencyrobot}
H.~Bharadhwaj, J.~Vakil, M.~Sharma, A.~Gupta, S.~Tulsiani, and V.~Kumar.
\newblock {RoboAgent}: Generalization and efficiency in robot manipulation via semantic augmentations and action chunking.
\newblock In \emph{Int. Conf. on Robotics and Automation ({ICRA})}, 2024.

\bibitem[Tan and Le(2019)]{Tan2019EfficientNetRM}
M.~Tan and Q.~Le.
\newblock {E}fficientnet: {R}ethinking model scaling for convolutional neural networks.
\newblock In \emph{Int. Conf. on Machine Learning ({ICML})}, 2019.

\bibitem[Kim et~al.(2025)Kim, Sim, Kim, Sycara, and Nam]{care}
J.~Kim, J.~Sim, W.~Kim, K.~P. Sycara, and C.~Nam.
\newblock {CARE}: {E}nhancing {S}afety of {V}isual {N}avigation through {C}ollision {A}voidance via {R}epulsive {E}stimation.
\newblock In \emph{Conf. on Robot Learning ({CoRL})}, 2025.

\bibitem[Zeng et~al.(2025)Zeng, Ren, Wang, Huang, and Cheng]{zeng2025navidiffusorcostguideddiffusionmodel}
Y.~Zeng, H.~Ren, S.~Wang, J.~Huang, and H.~Cheng.
\newblock {N}avidiffusor: {C}ost-guided diffusion model for visual navigation.
\newblock In \emph{Int. Conf. on Robotics and Automation ({ICRA})}, 2025.

\bibitem[Cai et~al.(2025)Cai, Peng, Yang, Zhang, Wei, Wang, Chen, Wang, and Pang]{cai2025navdplearningsimtorealnavigation}
W.~Cai, J.~Peng, Y.~Yang, Y.~Zhang, M.~Wei, H.~Wang, Y.~Chen, T.~Wang, and J.~Pang.
\newblock Navdp: Learning sim-to-real navigation diffusion policy with privileged information guidance, 2025.
\newblock URL \url{https://arxiv.org/abs/2505.08712}.

\bibitem[Siméoni et~al.(2025)Siméoni, Vo, Seitzer, Baldassarre, Oquab, Jose, Khalidov, Szafraniec, Yi, Ramamonjisoa, Massa, Haziza, Wehrstedt, Wang, Darcet, Moutakanni, Sentana, Roberts, Vedaldi, Tolan, Brandt, Couprie, Mairal, Jégou, Labatut, and Bojanowski]{dinov3}
O.~Siméoni, H.~V. Vo, M.~Seitzer, F.~Baldassarre, M.~Oquab, C.~Jose, V.~Khalidov, M.~Szafraniec, S.~Yi, M.~Ramamonjisoa, F.~Massa, D.~Haziza, L.~Wehrstedt, J.~Wang, T.~Darcet, T.~Moutakanni, L.~Sentana, C.~Roberts, A.~Vedaldi, J.~Tolan, J.~Brandt, C.~Couprie, J.~Mairal, H.~Jégou, P.~Labatut, and P.~Bojanowski.
\newblock Dinov3, 2025.
\newblock URL \url{https://arxiv.org/abs/2508.10104}.

\bibitem[Oquab et~al.(2024)Oquab, Darcet, Moutakanni, Vo, Szafraniec, Khalidov, Fernandez, Haziza, Massa, El-Nouby, Assran, Ballas, Galuba, Howes, Huang, Li, Misra, Rabbat, Sharma, Synnaeve, Xu, Jegou, Mairal, Labatut, Joulin, and Bojanowski]{oquab2024dinov2learningrobustvisual}
M.~Oquab, T.~Darcet, T.~Moutakanni, H.~Vo, M.~Szafraniec, V.~Khalidov, P.~Fernandez, D.~Haziza, F.~Massa, A.~El-Nouby, M.~Assran, N.~Ballas, W.~Galuba, R.~Howes, P.-Y. Huang, S.-W. Li, I.~Misra, M.~Rabbat, V.~Sharma, G.~Synnaeve, H.~Xu, H.~Jegou, J.~Mairal, P.~Labatut, A.~Joulin, and P.~Bojanowski.
\newblock Dinov2: Learning robust visual features without supervision, 2024.
\newblock URL \url{https://arxiv.org/abs/2304.07193}.

\bibitem[Hirose et~al.(2019)Hirose, Xia, Mart{\'i}n-Mart{\'i}n, Sadeghian, and Savarese]{go_stanford}
N.~Hirose, F.~Xia, R.~Mart{\'i}n-Mart{\'i}n, A.~Sadeghian, and S.~Savarese.
\newblock {D}eep {V}isual {MPC}-{P}olicy {L}earning for {N}avigation.
\newblock \emph{Robotics and Automation Letters}, 2019.

\bibitem[Shah et~al.(2021)Shah, Eysenbach, Rhinehart, and Levine]{recon}
D.~Shah, B.~Eysenbach, N.~Rhinehart, and S.~Levine.
\newblock {R}apid {E}xploration for {O}pen-{W}orld {N}avigation with {L}atent {G}oal {M}odels.
\newblock In \emph{Conf. on Robot Learning ({CoRL})}, 2021.

\bibitem[Hirose et~al.(2024)Hirose, Shah, Sridhar, and Levine]{hirose2023sacson}
N.~Hirose, D.~Shah, A.~Sridhar, and S.~Levine.
\newblock {SACSoN}: {S}calable {A}utonomous {C}ontrol for {S}ocial {N}avigation.
\newblock \emph{Robotics and Automation Letters}, 2024.
\newblock \doi{10.1109/LRA.2023.3329626}.

\bibitem[Karnan et~al.(2022)Karnan, Nair, Xiao, Warnell, Pirk, Toshev, Hart, Biswas, and Stone]{karnan2022scand}
H.~Karnan, A.~Nair, X.~Xiao, G.~Warnell, S.~Pirk, A.~Toshev, J.~Hart, J.~Biswas, and P.~Stone.
\newblock {S}ocially {C}ompli{A}nt {N}avigation {D}ataset ({SCAND}): {A} {L}arge-{S}cale {D}ataset of {D}emonstrations {F}or {S}ocial {N}avigation.
\newblock \emph{IEEE Robotics and Automation Letters}, 2022.

\bibitem[Urmson et~al.(2007)Urmson, Bagnell, Baker, Hebert, Kelly, Rajkumar, Rybski, Scherer, Simmons, Singh, et~al.]{urmson2007tartan}
C.~Urmson, J.~A. Bagnell, C.~Baker, M.~Hebert, A.~Kelly, R.~Rajkumar, P.~E. Rybski, S.~Scherer, R.~Simmons, S.~Singh, et~al.
\newblock {T}artan racing: A multi-modal approach to the {DARPA} urban challenge.
\newblock 2007.

\bibitem[Zhao et~al.(2021)Zhao, Zhang, Wang, Nogueira, and Scherer]{zhao2021super}
S.~Zhao, H.~Zhang, P.~Wang, L.~Nogueira, and S.~Scherer.
\newblock Super odometry: {IMU}-centric {LiDAR}-visual-inertial estimator for challenging environments.
\newblock In \emph{Int. Conf. on Intel. Robots and Sys. ({IROS})}, 2021.

\end{thebibliography}
\end{document}